\documentclass[10pt,a4paper,onecolumn,oneside]{article}

\usepackage{graphicx}
\usepackage{amsmath}
\usepackage{verbatim}
\usepackage{amssymb}
\usepackage{linguex}
\usepackage[x11names,dvipsnames]{xcolor}
\usepackage{tikz}
\usetikzlibrary{arrows,automata}
\usepackage{booktabs}
\usepackage{multirow}
\usepackage[english]{babel}
\usepackage[utf8x]{inputenc}
\usepackage[f]{esvect}
\usepackage{authblk}
\usepackage{pxfonts}
\usepackage{xspace}
\usepackage{soul}
\usepackage{centernot}
\usepackage[nodayofweek]{datetime}
\usepackage{enumitem}
\usepackage[T1]{fontenc}
\usepackage{hyperref}

\setlist[enumerate,1]{label=\textbf{\textit{(\roman*)}}, topsep=0.2em, itemsep=0.2em, leftmargin=2em}
\setlist[itemize,1]{label=\textbullet, topsep=0.35em, itemsep=0.2em, leftmargin=2em}

\newlist{compactitemize}{itemize}{1}
\setlist[compactitemize,1]{label=\textbullet, leftmargin=1em, rightmargin=0em, topsep=0.25em, itemsep=0em}

\newlist{compactenumerate}{enumerate}{1}
\setlist[compactenumerate,1]{label=\textbf{\textit{(\roman*)}}, leftmargin=1.5em, rightmargin=0em, topsep=0.25em, itemsep=0.25em}

\newlist{inlineenum}{enumerate*}{1}
\setlist[inlineenum,1]{label=\textbf{\textit{(\roman*)}}}

\newlist{inductionproof}{description}{1}
\setlist[inductionproof,1]{topsep=0.4em, leftmargin=0em, itemsep=0.25em}

\newcommand{\colornew}{Bittersweet}

\newtheorem{Def}{Definition}
\newtheorem{Theorem}{Theorem}
\newtheorem{Prop}{Proposition}

\newtheorem{Lemma}{Lemma}
\newtheorem{exa}{Example}

\newenvironment{ourproof}
{\noindent\textit{Proof.}}
{\hfill\qed\medskip}

\newcommand{\qed}{$\blacksquare$}

\renewcommand{\arraystretch}{1.3}

\renewcommand{\S}{\mathcal{S}}
\newcommand{\NV}{\mathcal{V}}
\newcommand{\XV}{\mathcal{U}}
\newcommand{\Ran}{\mathcal{R}}
\newcommand{\R}[1]{\Ran(#1)}
\newcommand{\F}{\mathcal{F}}
\newcommand{\fun}{f}
\newcommand{\A}{\mathcal{A}}
\newcommand{\B}{\mathcal{B}}
\renewcommand{\vec}[1]{\vv{#1}}
\newcommand{\ina}{{=}}
\newcommand{\opint}[2]{#1\ina#2}
\newcommand{\newval}[3]{{{#1}^{#2}_{#3}}}
\newcommand{\T}{\mathcal{T}}
\newcommand{\ec}{\operatorname{E}}
\newcommand{\LA}{\ensuremath{\mathcal{L}}\xspace}
\newcommand{\LAfull}{{\ensuremath{\mathcal{L}_{\mathit{PAKC}}}}\xspace}
\newcommand{\LAkc}{\ensuremath{\mathcal{L}_{\mathit{KC}}}\xspace}
\newcommand{\LAc}{\ensuremath{\mathcal{L}_{\mathit{C}}}\xspace}
\newcommand{\LAcod}{\ensuremath{\mathcal{L}_{\mathit{COD}}}\xspace}
\newcommand{\LOfull}{\ensuremath{\mathsf{L}_{\mathit{PAKC}}}\xspace}
\newcommand{\LOkc}{\ensuremath{\mathsf{L}_{\mathit{KC}}}\xspace}

\newcommand{\tro}{\operatorname{tr_1}}
\newcommand{\trt}{\operatorname{tr_2}}
\newcommand{\mcss}{\mathbb{C}}
\newcommand{\rmcss}{\mathbb{D}}
\newcommand{\dep}[2]{=\hspace{-3pt}({#1};{#2})}

\newcommand{\parentof}{\hookrightarrow}
\newcommand{\synparentof}{\rightsquigarrow}

\renewcommand{\int}[1]{[#1]}

\newcommand{\bs}[1]{\boldsymbol{#1}}
\newcommand{\set}[1]{\{ #1 \}}
\newcommand{\tuple}[1]{\langle #1 \rangle}
\newcommand{\card}[1]{\mathopen{\vert} #1 \mathclose{\vert}}
\newcommand{\Nat}{\mathbb{N}}
\newcommand{\prooflr}{\ensuremath{\bs{(\Rightarrow)}}\xspace}
\newcommand{\proofrl}{\ensuremath{\bs{(\Leftarrow)}}\xspace}

\newenvironment{ctabular}[1]
{\begin{center}\begin{tabular}{#1}}
{\end{tabular}\end{center}}

\newenvironment{footnotesizectabular}[1]
{\begin{center}\begin{footnotesize}\begin{tabular}[b]{#1}}
{\end{tabular} \end{footnotesize}\end{center}}

\newenvironment{smallctabular}[1]
{\begin{center}\begin{small}\begin{tabular}[b]{#1}}
{\end{tabular}\end{small}\end{center}}

\newcommand{\keywords}[1]{\medskip\noindent\textbf{Keywords}: #1}

\begin{document}

\title{\Large \bf Thinking About Causation:\\ A Causal Language with Epistemic Operators\thanks{\texttt{\jobname.tex}, \emph{compiled \today, \currenttime.}}}

\author[1]{\small Fausto Barbero}
\author[2]{Katrin Schulz}
\author[2,3]{Sonja Smets}
\author[2]{Fernando R. Vel{\'a}zquez-Quesada}
\author[2]{Kaibo Xie}

\affil[1]{\small University of Helsinki. \texttt{fausto.barbero@helsinki.fi}}
\affil[2]{\small ILLC, Universiteit van Amsterdam. \texttt{\{K.Schulz,S.J.L.Smets@uva.nl,F.R.VelazquezQuesada,K.Xie\}@uva.nl}}
\affil[3]{\small Department of Information Science and Media Studies, University of Bergen.}

\date{}

\maketitle

\begin{abstract}
  This paper proposes a formal framework for modeling the interaction of causal and (qualitative) epistemic reasoning. To this purpose, we extend the notion of a causal model \cite{Galles,halpern2000axiomatizing,Pearl09,briggs2012interventionist} with a representation of the epistemic state of an agent. On the side of the object language, we add operators to express knowledge and the act of observing new information. We provide a sound and complete axiomatization of the logic, and discuss the relation of this framework to causal team semantics. 

\end{abstract}
%
%
%


\section{Introduction}\label{sec:introduction}

In recent years a lot of effort has been put in the development of formal models of causal reasoning. A central motivation behind this is the importance of causal reasoning for AI. Making computers take into account causal information is currently one of the central challenges of AI research \cite{PearlWhy,Bergstein}. There has also been tremendous progress in this direction after the earlier groundbreaking work in \cite{Pearl00} and \cite{Spirtes}. Advanced formal and computational tools have been developed for modelling causal reasoning and learning causal information, with applications 
in many different scientific areas. 
In this paper we want to extend this work further. The direction we want to explore is that of developing formal models of the interaction between causal and epistemic reasoning.

Even though the standard logical approach to causal reasoning (\cite{Pearl00,halpern2000axiomatizing,halpern2016actual}) can model epistemic uncertainty\footnote{E.g., by adding a probability distribution over a causal model's exogenous variables.}, it does not permit reasoning about the interaction between causal and epistemic reasoning in the object language. Although recently there have been proposals adding probabilistic expressions to the object language (e.g., \cite{ibeling2020probabilistic}), very little has been done on combining causal and qualitative epistemic reasoning.\footnote{See \cite{barbero2019interventionist} for an exception, though the epistemic element is not made fully explicit in the language. Section~\ref{sec:discussion} discusses the relationship between the referred paper and the current proposal.} However, this kind of reasoning occurs frequently in our daily life, especially in connection with counterfactual thinking. Consider, for instance, the following situation.

\begin{exa}\label{circuit}
  In front of Billie there is a button, which is connected to a circuit breaker and a sprinkler. If the circuit is closed, the sprinkler works if and only if the button is pushed. If the circuit is not closed, the sprinkler won't work, independently of the state of the button. Billie knows these causal laws. She can also see the button and the sprinkler, but she does not know the state of the circuit breaker. Suppose that at the moment the circuit is closed and the button is not pushed; as a result, the sprinkler is not working.
\end{exa}

In such a situation, we want to derive that Billie is not sure that if the button had been pushed, the sprinkler would have been working. 
Thus, we want to make inferences involving epistemic attitudes towards counterfactuals, which in turn explore causal dependencies. We also want to reason counterfactually about such epistemic attitudes. Considering the same example, we also want to infer that if Billie had pushed the button and saw that the sprinkler works, then she would have known that the circuit is closed (because of the causal knowledge she has). In order to formalize this type of reasoning, we need a framework that combines causal reasoning with a model of epistemic attitudes.

{\smallskip}

Given the vast literature on epistemic logic, there is a lot of work that we can build on. This paper makes a start on combining the standard approach to causal reasoning (\cite{Pearl00,halpern2000axiomatizing,halpern2016actual}) with tools from Dynamic Epistemic Logic (DEL; \cite{BaltagMossSolecki1998,vanBenthem2011ldii,vanDitmarschEtAl2007}). The main motivation for this choice is the dynamic character of both systems, even though this aspect will not be explored at depth here. For now we will only consider a very simple extension of the standard system of causal reasoning. But, as we will show, this basic extension already allows us to formalise some interesting concepts and formulate concrete questions for further research.

\medskip

\noindent\textbf{Outline}. Section~\ref{sec:standard} introduces the standard approach to causal reasoning, and then Section~\ref{sec:motivation} motivates in more detail the extension proposed here. Section~\ref{sec:proposal} extends the standard causal modeling with means to express knowledge and external communication, and Section~\ref{sec:axiomatisation} provides a sound and complete axiomatization for the new system. Section~\ref{sec:discussion} concludes the paper discussing the relationship with Causal Team Semantics \cite{barbero2019interventionist,BarSan2020}. 

\section{The standard causal modelling approach}\label{sec:standard}

What we refer to as the standard logic of causal reasoning was presented on \cite{Pearl95}, extended in \cite{Galles}, and then further developed in, among others, \cite{halpern2000axiomatizing,pearl2002causality,briggs2012interventionist}. This section recall briefly the most important concepts and tools.

{\medskip}

The starting point is a formal representation of causal dependencies. This is done in terms of causal models, which represent the causal relationships between a finite set of  variables. These variables as well as their ranges of values are given by a \emph{signature}. Throughout this text, let $\S = \tuple{\XV,\NV,\Ran}$ be the \emph{finite} signature where
\begin{itemize}
  \item $\XV = \set{U_1, \dots, U_m}$ is a finite set of \emph{exogenous} variables (those whose value is causally independent from the value of every other variable in the system),
  \item $\NV = \set{V_1, \ldots, V_n}$ is a finite set of \emph{endogenous} variables (those whose value is completely determined by the value of other variables in the system), and
  \item $\R{X}$ is the finite non-empty range of the variable $X \in \XV \cup \NV$.\footnote{Given $(X_1,\ldots,X_k) \in (\XV \cup \NV)^k$, abbreviate $\R{X_1}\times \cdots \times\R{X_k}$ as $\R{X_1,\ldots,X_k}$.}
\end{itemize}


A causal model is formally defined as follows.

\begin{Def}[Causal model]\label{CausalModel}
  A \emph{causal model} is a triple $\tuple{\S, \F, \A}$ where
  \begin{itemize}
    \item $\S = \tuple{\XV,\NV,\Ran}$ is the model's signature,
    \item $\F = \set{ \fun_{V_j} \mid V_j \in \NV }$ assigns, to each endogenous variable $V_j$, a map
    \[
      \fun_{V_j}:\R{U_1, \ldots, U_m, V_1, \ldots, V_{j-1}, V_{j+1}, \ldots, V_n} \to \R{V_j}.
    \]
    The map $\fun_V$ is sometimes called $V$'s \emph{structural function}, and the set $\F$ is called a set of structural functions for $\NV$.

    \item $\A$ is the \emph{valuation} function, assigning to every $X \in \XV \cup \NV$ a value $\A(X) \in \R{X}$. For each endogenous variable, the valuation should \emph{comply with} the variable's structural function. In other words, for every $V_j \in \NV$, the following should hold:
    \[ \A(V_j) = \fun_{V_j} \left( \A(U_1), \ldots,\A(U_m),\A(V_1),\ldots,\A(V_{j-1}),\A(V_{j+1}),\ldots,\A(V_n) \right). \]
  \end{itemize}
\end{Def}

In a causal model $\tuple{\S, \F, \A}$, the functions in $\F$ describe the causal relationship between the variables. Using these functional dependencies, we can define what it means for a variable to directly causally affect another variable.\footnote{This notion of a {\it direct cause} is adopted from \cite{Galles}; it is related to the notion of a variable having a {\it direct effect} on another, as discussed in \cite{Pearl00} in the context of Causal Bayes Nets. The notions defined here differ from Halpern's notion of {\it affect} \cite{halpern2000axiomatizing}, and this affects the axiomatization: axiom HP6 (Table \ref{tbl:axioms}) has the same function as C6 in \cite{halpern2000axiomatizing} (ensuring that the canonical model is recursive), but does so in a slightly different way.}

\begin{Def}[Causal dependency]\label{def:parentof}
   Let $\F$ be a set of structural functions for $\NV$. Given an endogenous variable $V_j \in \NV$, rename each other variable in $\S$, the variables $U_1, \ldots, U_m, V_1, \ldots, V_{j-1}, V_{j+1}$, $\ldots, V_n$, as $X_1, \ldots, X_{m+n-1}$, respectively.

   We say that, under the structural functions in $\F$, an endogenous variable $V_j \in \NV$ is \emph{directly causally affected} by a variable $X_i \in (\XV \cup \NV) \setminus \set{V_j}$ (in symbols, $X_i \parentof_\F V_j$) if and only if there is a tuple
   \[
     (x_1, \ldots, x_{i-1}, x_{i+1}, \ldots, x_{m+n-1})
     \in
     \R{X_1, \ldots, X_{i-1}, X_{i+1}, \ldots, X_{m+n-1}}
  \]
   and there are $x'_i \neq x''_i \in \R{X_i}$ such that
   \[ \fun_{V_j} (x_1, \ldots, x'_{i}, \ldots, x_{m+n-1}) \neq \fun_{V_j} (x_1, \ldots, x''_{i}, \ldots, x_{m+n-1}). \]
   When $X_i \parentof_\F V_j$, we will also say that $X_i$ is a \emph{causal parent} of $V_j$.
    The relation 
    $\parentof_\F^+$ is the transitive closure of $\parentof_\F$.
\end{Def}

As it is common in the literature, we restrict ourselves to causal models in which circular causal dependencies do not occur.\footnote{The reason behind this restriction is that only acyclic relations are thought to have a causal interpretation (see \cite{StrWol1960} for an argument). The counterfactuals satisfy different logical laws if cyclic dependencies are allowed (see \cite{halpern2000axiomatizing}).}

\begin{Def}[Recursive causal model]
  A set of structural functions $\F$ is \emph{recursive} if and only if $\parentof_\F^+$ is a strict partial order (i.e., an asymmetric [hence irreflexive] and transitive relation, so there are no cycles). A causal model $\tuple{\S, \F, \A}$ is \emph{recursive} if and only if $\F$ is recursive. In this text, a recursive causal model will be called simply a \emph{causal model}.
\end{Def}

The most important notion of this formalisation of causal reasoning is that of an \emph{intervention}. This notion refers to the action of changing the values of variables in the system. Before we define an intervention formally, let us first introduce the notion of assignment.

\begin{Def}[Assignment]
  Let $\S = \tuple{\XV,\NV,\Ran}$ be a signature. An \emph{assignment} on $\S$ is an expression $\opint{\vec{X}}{\vec{x}}$ where $\vec{X}$ is a tuple of different variables in $\XV \cup \NV$ (that is, $\vec{X} = (X_1, \ldots, X_k) \in (\XV\cup\NV)^k$ for some $k \in \Nat$, with $X_i \neq X_j$ for $i \neq j$), and $\vec{x} \in \R{\vec{X}}$.
\end{Def}

Now, an intervention that sets a variable $X$ to the value $x$ is defined as an operation that maps a given model $M$ to a new model $M_{X=x}$, which is the same as $M$ except that the function determining the value of $X$ is replaced by the constant function mapping $X$ to $x$. In other words, $X$ is cut off from all its causal dependencies and fixed to the value $x$.

\begin{Def}[Intervention]\label{def:int:strict}
  Let $M = \tuple{\S, \F, \A}$ be a causal model; let $\opint{\vec{X}}{\vec{x}}$ be an assignment on $\S$. The causal model $M_{\opint{\vec{X}}{\vec{x}}} = \tuple{\S, \F_{\opint{\vec{X}}{\vec{x}}}, \newval{\A}{\F}{\opint{\vec{X}}{\vec{x}}}}$, resulting from an intervention setting the values of variables in $\vec{X}$ to $\vec{x}$, is such that
  \begin{itemize}
    \item $\F_{\opint{\vec{X}}{\vec{x}}}$ is as $\F$ except that, for each endogenous variable $X_i$ in $\vec{X}$, the function $\fun_{X_i}$ is replaced by a \emph{constant} function $\fun'_{X_i}$ that returns the value $x_i$ regardless of the values of all other variables.

    \item $\newval{\A}{\F}{\opint{\vec{X}}{\vec{x}}}$ is the unique valuation where \begin{inlineenum} \item the value of each exogenous variable not in $\vec{X}$ is exactly as in $\A$, \item the value of each each exogenous variable $X_i$ in $\vec{X}$ is the provided $x_i$, and \item the value of each endogenous variable complies with its \emph{new} structural function (that in $\F_{\opint{\vec{X}}{\vec{x}}}$)\end{inlineenum}.\footnote{Note that, since $\F$ is recursive, the valuation $\newval{\A}{\F}{\opint{\vec{X}}{\vec{x}}}$ is uniquely determined. First, the value of every exogenous variable $U$ is uniquely determined, either from $\vec{x}$ (if $U$ occurs in $\vec{X}$) or else from $\A$ (if $U$ does not occur in $\vec{X}$). Second, the value of every endogenous variable $V$ is also uniquely determined, either from $\vec{x}$ (if $V$ occurs in $\vec{X}$, as $V$'s new structural function is a constant) or else from the (recall: recursive) structural functions in $\F_{\opint{\vec{X}}{\vec{x}}}$ (if $V$ does not occur in $\vec{X}$).}
  \end{itemize}
\end{Def}

We can now extend a propositional language with a new type of sentence for describing the effect of an intervention. The expression $\int{\opint{\vec{X}}{\vec{x}}}\gamma$ should be read as the counterfactual conditional {\it if the variables in $\vec{X}$ were set to the values $\vec{x}$, respectively, then $\gamma$ would be the case}.

\begin{Def}
  Formulas $\phi$ of the language \LAc based on the signature $\S$ are given by
  \[
    \begin{array}{@{}l@{\;::=\;}l@{\;\quad\;}l@{}}
      \gamma & Z{=}z \mid \lnot \gamma \mid \gamma \land \gamma
             & \text{for } Z \in \XV \cup \NV \text{ and } z \in \R{Z} \\
      \phi   & Z{=}z \mid \lnot \phi \mid \phi \land \phi \mid \int{\opint{\vec{X}}{\vec{x}}}\gamma
             & \text{for }  \opint{\vec{X}}{\vec{x}} \text{ an assignment on } \S \\
    \end{array}
  \]
\end{Def}

The language makes free use of Boolean operators, but it forbids the nesting of intervention operators $\int{\opint{\vec{X}}{\vec{x}}}$ (see \cite{briggs2012interventionist} for a way to remove this restriction). Formulas of \LAc are evaluated in causal models $\tuple{\S, \F, \A}$. The semantic interpretation for Boolean operators is the usual; for the rest,
\begin{ctabular}{l@{\qquad{iff}\qquad}l}
  $\tuple{\S, \F, \A} \models Z{=}z$ & $\A(Z) = z$ \\
  $\tuple{\S, \F, \A} \models \int{\opint{\vec{X}}{\vec{x}}}\gamma$ & $\tuple{\S, \F_{\opint{\vec{X}}{\vec{x}}}, \newval{\A}{\F}{\opint{\vec{X}}{\vec{x}}}} \models \gamma$ \\
\end{ctabular}



\section{Limitations of the standard system}\label{sec:motivation}

The notion of a causal model contains an incredible amount of extra 
information compared to classical models. 
Not only does it tell us which variables depend causally on which other variables, but it also determines the exact character of this dependence. On the side of the language this wealth of information is then explored in terms of counterfactual conditionals using the concept of an intervention. This is where the actual causal reasoning happens. The standard logic of causal reasoning is in fact a logic of counterfactual reasoning. This is no accident: Judea Pearl, founder of the approach to causal reasoning introduced above, sees both concepts as intimately related. He argues that only when an agent can evaluate counterfactual conditionals does she fully engage with causal reasoning \cite{Pearl09,PearlWhy}. Counterfactual reasoning {\it is} the highest level of causal reasoning -- a level that even the most advanced AI technology doesn't reach.\footnote{The other two levels that Pearl distinguishes are the level of association, which is based on observation, and the level of intervention, which is based on doing. Modern AI technology is for him still at the first level: association. Counterfactual reasoning is not possible without a true understanding of {\it why} things happen -- in our terminology, it is not possible without knowing the causal relationships as determined by $\F$.} 

Still, the basic causal framework has some limitations. An important one is that causal (or counterfactual) reasoning does not stand on its own: it does interact with other forms of reasoning. For instance, and as we illustrated in the introduction, counterfactual reasoning also considers the effect interventions have on the epistemic state of (observing) agents. We can reason that {\it If Peter had pushed the button, he would have known that his flashlight is broken}, which involves thinking about Peter's epistemic state after observing a causal intervention. This type of reasoning allows us to plan our actions (try out a flashlight before we take it for a night walk), and also influences our interaction with other agents (if you want Peter to come back from his walk, you should tell him to test his flashlight before he leaves). Therefore, a full account of the logic of causal reasoning needs to model its interaction with epistemic reasoning as well. The next section takes a first step in this direction: it adds a representation of the epistemic state of an agent to the model, extending the language with expressions that can talk about knowledge and knowledge-update in the context of causal reasoning.

There is another perspective from which such an epistemic extension of the standard framework can be motivated. In recent years there has been growing interest in the logic of dependence/determinacy. 
For instance, the IF logic of \cite{ManSanSev2011} expresses dependence by decorations of the quantifiers. Then, \cite{Vaa2007} and \cite{baltag2020simple} use a primitive expression indicating that the value of one variable depends on that of another. In all these cases, the discussed notion of dependence/determinacy relies on considering a multiplicity of valuations in the model: the variable $Y$ depends on (it is determined by) the variables $X_1, \ldots, X_n$ when, in all valuations that are being considered, fixing the value of the latter also fixes the value of the former. This gives rise to the question of how the notion of causal dependence modelled by the just introduced framework interacts with the notions of dependence/determinacy modelled by these alternative frameworks, and how causal dependence fits into a general picture of reasoning with and about dependencies. Interestingly, extending the standard causal reasoning approach with basic epistemic notions gives us another way to express the same notion of dependence as studied in the works just cited. This, then, allows us to compare different notions of dependency within one logical system. We will come back to this connection in Section~\ref{sec:discussion}.

\section{Epistemic causal models}\label{sec:proposal}

The first step towards a framework that combines causal with epistemic reasoning is adding a representation of the epistemic state of an agent to the causal model. This is done by adding a set of valuations $\T$, representing the alternatives the agent considers possible.

\begin{Def}[Epistemic causal model]
  An \emph{epistemic (note: recursive) causal model} is a tuple $\tuple{\S, \F, \T}$ where $\S = \tuple{\XV, \NV, \Ran}$ is a signature, $\F$ is a (note: recursive) set of structural functions for $\NV$, and $\T$ is a non-empty set of valuation functions for $\XV \cup \NV$, each one of them complying with $\F$.
\end{Def}

Example \ref{circuit} can now be modelled as follows. We define an epistemic causal model $\ec=\tuple{\S, \F, \T}$ whose signature $\S$ has three variables: the exogenous $B$ for the button and $C$ for the circuit breaker, and the endogenous $S$ for the sprinkler. All three variables can take two values, $0$ or $1$. The set of functions $\F$ contains only  one element: the function mapping $S$ to $1$ iff both $B$ and $C$ also have value $1$. Because the agent can observe the value of the variables $B$ and $S$, the set $\T$ contains the assignment $\A_1$ that maps $C$ to $0$, $B$ to $0$ and $S$ to $0$, and the assignment $\A_2$ that maps $C$ to $1$, $B$ to $0$ and $S$ to $0$. Note how $\T$ cannot contain the assignment $C=1$, $B=1$ and $S=0$, for instance, because this assignment does not comply  with the causal law in $\F$.
This observation highlights an important feature of this notion of epistemic model: it cannot model uncertainty about the causal dependencies. Investigating the consequences of lifting this restriction is left for future research.

{\smallskip}

The next step is to extend the notion of intervention to epistemic causal models.

\begin{Def}[Intervention]\label{def:intervention2}
  Let $\ec = \tuple{\S, \F, \T}$ be an epistemic causal model; let $\opint{\vec{X}}{\vec{x}}$ be an assignment on $\S$. The epistemic causal model $\ec_{\opint{\vec{X}}{\vec{x}}} = \tuple{\S, \F_{\opint{\vec{X}}{\vec{x}}}, \newval{\T}{\F}{\opint{\vec{X}}{\vec{x}}}}$, resulting from an intervention setting the values of variables in $\vec{X}$ to $\vec{x}$, is such that
  \begin{itemize}
    \item $\F_{\opint{\vec{X}}{\vec{x}}}$ is defined from $\F$ just as in Definition \ref{def:int:strict},

    \item $\newval{\T}{\F}{\opint{\vec{X}}{\vec{x}}} := \set{ \newval{\A'}{\F}{\opint{\vec{X}}{\vec{x}}} \mid \A' \in \T}$ (see Definition \ref{def:int:strict}).
  \end{itemize}
  Note how $\tuple{\S, \F_{\opint{\vec{X}}{\vec{x}}}, \newval{\T}{\F}{\opint{\vec{X}}{\vec{x}}}}$ is indeed an epistemic causal model, as $\F_{\opint{\vec{X}}{\vec{x}}}$ is recursive and all valuations in $\newval{\T}{\F}{\opint{\vec{X}}{\vec{x}}}$ comply with it.
\end{Def}

In the just introduced model $\ec$ for Example~\ref{circuit}, we can now calculate the effects of considering the intervention that sets $B=1$. According to Definition~\ref{def:intervention2}, an intervention on an epistemic causal model amounts to intervening on each of the assignments contained in the epistemic state. Thus, for our concrete example, we need to calculate the effects of an intervention with $B=1$ on the assignments $\A_1$ and $\A_2$ that make up the epistemic state $\T$. The new epistemic state $\T^{\F}_{B=1}$ will now contain the assignment $\A^{\F}_{1, B=1}$ that maps $C$ to $0$, $B$ to $1$ and $S$ to $0$ and the assignment $\A^{\F}_{2, B=1}$ that maps $C$ to $1$, $B$ to $1$ and $S$ to $1$. Thus, the consequences of the intervention are calculated for all epistemic possibilities the agent considers. In other words, Definition~\ref{def:intervention2} assumes that the agent has full epistemic access to the effect of the intervention on the model. In particular, she knows that the intervention takes place (in the counterfactual scenario considered). This makes a lot of sense if you think of the agent whose epistemic state is modelled as the one engaging in the counterfactual thinking. It is less plausible in connection to counterfactual thinking about the knowledge states of other agents. But this is something that we can leave for now, as we will not consider epistemic causal models for multiple agents in this paper.

{\smallskip}

Based on these changes on the semantic side, we can now extend the object language with expressions that talk about the epistemic state of the agent. More specifically, we add the operator $K$ for knowledge and ``$!$'' for information update. In other words, we understand ``$!$'' as expressing the action of observing or receiving information.

\begin{Def}
  Formulas $\phi$ of the language \LAfull based on $\S$ are given by
  \[
    \begin{array}{l@{\;::=\;}l@{\quad}l}
      \gamma & Z{=}z \mid \lnot \gamma \mid \gamma \land \gamma \mid K\gamma \mid [\gamma!]\gamma
             & \text{for } Z \in \XV \cup \NV \text{ and } z \in \R{Z} \\
      \phi   & Z{=}z \mid \lnot \phi \mid \phi \land \phi \mid K\phi \mid [\phi!]\phi \mid \int{\opint{\vec{X}}{\vec{x}}}\gamma
             & \text{for } \opint{\vec{X}}{\vec{x}} \text{ an assignment on } \S
    \end{array}
  \]
\end{Def}

Other Boolean operators ($\vee, \rightarrow, \leftrightarrow$) can be defined as usual. Note how, although the language makes free use of Boolean, epistemic and announcement operators ($K$ and $[\phi!]$, for the latter two), nested intervention is again not allowed.\footnote{However, notice that the semantics already allows for nested occurrences of all dynamic operators. We will extend the proofs of sound- and completeness to the unrestricted language in the future.}
Note also how the tuple vector $\vec{X}$ can be empty, in which case $\int{\opint{\vec{X}}{\vec{x}}}\gamma$ becomes $\gamma$. The semantics for this extended language is straightforward.

\begin{Def}
  Formulas of \LAfull are evaluated in a pairs $(\ec, \A)$ with $\ec = \tuple{\S, \F, \T}$ an epistemic causal model and $\A \in \T$. The semantic interpretation for Boolean operators is the usual; for the rest,
  \begin{ctabular}{l@{\qquad{iff}\qquad}l}
    $(\ec, \A) \models Z{=}z$ & $\A(Z) = z$ \\
    $(\ec, \A) \models K\phi$ & $(\ec, \A') \models \phi$ for every $\A' \in \T$ \\
    $(\ec, \A) \models [\psi!]\phi$ & $(\ec, \A) \models \psi$ implies $(\ec^{\psi}, \A) \models \phi$ \\
    $(\ec, \A) \models \int{\opint{\vec{X}}{\vec{x}}}\gamma$ & $(\ec_{\opint{\vec{X}}{\vec{x}}}, \newval{\A}{\F}{\opint{\vec{X}}{\vec{x}}}) \models \gamma$ \\
  \end{ctabular}
  with $\ec^{\psi} = \tuple{\S, \F, \T^{\psi}}$ such that $\T^{\psi} := \set{\A' \in \T \mid (\ec, \A') \models \psi}$. Note how $\ec^{\psi}$ is an epistemic causal model: $\F$ is recursive, and all valuations in $\T^{\psi}$ comply with it.
\end{Def}


To illustrate this definition, we go back to the epistemic model $\ec$ introduced for Example~\ref{circuit}. In order to evaluate a concrete formula with respect to this model we need to select, next to $\ec$, an assignment representing the actual world. In the example this is assignment $\A_2$: in the actual world, the circuit breaker is closed, but because the button has not been pushed, the sprinkler is not working. We can calculate that the counterfactual $[B{=}1]S{=}1$ comes out as true given $\ec$ and $\A_2$, just as in the non-epistemic approach discussed in Section~\ref{sec:standard}. But because we now also have a representation of the epistemic state of some agent, we can additionally consider epistemic attitudes the agent has towards this counterfactual. For instance, we can check that $K([B{=}1]S{=}1)$ is not true given $\ec$ and $\A_2$. For the sentence to be true, the formula $[B{=}1]S{=}1$ needs to be true over both $(\ec, \A_1)$ and $(\ec, \A_2)$, because $\A_1$ and $\A_2$ are the two elements of $\T$. Thus, we need both $(\ec_{B{=}1}, \A^{\F}_{1, B{=}1}) \models S{=}1$ and $(\ec_{B{=}1}, \A^{\F}_{2, B{=}1}) \models S{=}1$. We already calculated $\T_{B{=}1}$ above: $\T_{B{=}1}=\{\A^{\F}_{1, B{=}1}, \A^{\F}_{2, B{=}1}\}$. But the content of $\T_{B{=}1}$ does not matter for the truth of the consequent $S{=}1$ of the counterfactual that we are considering here, since this consequent does not contain epistemic operators. However, while in $\A^\F_{1, B{=}1}$ the sprinkler is still off, in $\A^\F_{2, B{=}1}$ it is on. This means that $(\ec_{B{=}1}, \A^{\F}_{1, B{=}1}) \not \models S{=}1$, while $(\ec_{B{=}1}, \A^{\F}_{2, B{=}1}) \models S{=}1$. Thus, the agent cannot predict the outcome of the intervention, just as intended in this case.

{\smallskip}

Finally, we define an operator $\synparentof$ in terms of the existing vocabulary as a way to express causal dependency in the object language.

\begin{Def}\label{def:syntparentof}
  Take $X$ and $Z$ in $\XV \cup \NV$. The formula $X \synparentof Z$ is defined as
  \[
    \bigvee_{
      \renewcommand{\arraystretch}{1.2}
      \begin{array}{l}
        \vec{w} \in \R{(\XV \cup \NV) \setminus \set{X,Z}}, \\
        \set{x_1, x_2} \subseteq \R{X}, x_1 \neq x_2, \\
        \set{z_1, z_2} \subseteq \R{Z}, z_1 \neq z_2
      \end{array}
    }
    \int{\opint{\vec{W}}{\vec{w}},\opint{X}{x_1}}Z{=}z_1 \;\wedge\; \int{\opint{\vec{W}}{\vec{w}},\opint{X}{x_2}}Z{=}z_2,
  \]
\end{Def}


A formula $X \synparentof Z$ should be read as \emph{``$X$ has a direct causal effect on $Z$''}. It holds when there is a vector $\vec{w}$ of values for variables in $\R{\XV \cup \NV \setminus \set{X,V}}$ and two different values $x_1, x_2$ for $X$ that produce two different values $z_1, z_2$ for $Z$ (cf. \cite{halpern2000axiomatizing}). When $Z \in \NV$, it is clear that $\synparentof$ is the syntactic counterpart of the relation ``$\parentof$'' of Definition~\ref{def:parentof}.



\section{Axiomatization}\label{sec:axiomatisation}

The axiom system \LOfull is presented in Table \ref{tbl:axioms}.
The \textit{intervention} axioms, HP1-HP6, RH1 and RH2, 
are the standard axiomatization for the intervention operator over recursive causal models, with EX an additional axiom indicating that an exogenous variable is immune to interventions to any other variables. Then, the \textit{epistemic} part contains the standard modal S5 axiomatization for truthful knowledge with positive and negative introspection.

Axiom CM indicates that what the agent will know after an intervention ($\int{\opint{\vec{X}}{\vec{x}}}K\phi$) is exactly what she knows now about the effects of the intervention ($K\int{\opint{\vec{X}}{\vec{x}}}\phi$). Although maybe novel in the literature on causal models, the axiom is simply an instance of the more general DEL pattern of interaction between knowledge and a deterministic action without precondition. Finally, axioms RP2-RP4 and rule RE in the \textit{announcement} part are a \emph{reduction-based} axiomatisation for public announcements in the DEL style. Here, axioms RP4 and RP1 are the most important. The first, RP4, is the well-known reduction axiom for announcement and knowledge, stating that knowing $\phi$ after an announcement of $\psi$ is equivalent to knowing, conditionally on $\psi$, that the announcement of $\psi$ would make $\phi$ true.\footnote{Note how the announcement of $\psi$ is a deterministic action \emph{with precondition $\psi$}. Hence the similarities and differences between RP4 and CM.} The second, RP1, establishes the reduction for `atoms' of the form $\int{\opint{\vec{X}}{\vec{x}}}Z{=}z$; when $\vec{X}$ is not empty, it states that a public announcement does not change the causal rules in the model.

\begin{table}[!ht]
  \renewcommand{\arraystretch}{1.5}
  \begin{footnotesizectabular}{r@{:\;\,}l@{\qquad}r@{:\;\,}l}
    \toprule
    \multicolumn{4}{l}{\textit{Propositional}:} \\
    P & $\vdash \phi$ \;\; for $\phi$ an instance of a tautology &
    MP & From $\phi \rightarrow \psi$ and $\phi$ derive $\psi$ \\
    \midrule
    \multicolumn{4}{l}{\textit{Intervention}:} \\
    HP1 & \multicolumn{3}{@{}l}{$\vdash \int{\opint{\vec{X}}{\vec{x}}}Z{=}z \,\rightarrow\, \neg\int{\opint{\vec{X}}{\vec{x}}}Z{=}z'$ \qquad for $z \neq z' \in \R{Z}$} \\
    HP2 & \multicolumn{3}{@{}l}{$\vdash \bigvee_{z \in \R{Z}}\int{\opint{\vec{X}}{\vec{x}}}Z{=}z$} \\
    HP3 & \multicolumn{3}{@{}l}{$\vdash \left( \int{\opint{\vec{X}}{\vec{x}}}Z{=}z \,\wedge\, \int{\opint{\vec{X}}{\vec{x}}}W{=}w \right) \,\rightarrow\, \int{\opint{\vec{X}}{\vec{x}},\opint{Z}{z}}W{=}w$} \\
    HP4 & \multicolumn{3}{@{}l}{$\vdash \int{\opint{\vec{X}}{\vec{x}},\opint{Z}{z}}Z{=}z$} \\
    HP5 & \multicolumn{3}{@{}l}{$\vdash \left(\int{\opint{\vec{X}}{\vec{x}},\opint{Z}{z}}W{=}w \,\wedge\, \int{\opint{\vec{X}}{\vec{x}},\opint{W}{w}}Z{=}z\right) \,\rightarrow\, \int{\opint{\vec{X}}{\vec{x}}}W{=}w$ \qquad for $W \neq Z$} \\
    HP6 & \multicolumn{3}{@{}l}{$\vdash (Z_0 \synparentof Z_1 \;\wedge\; \cdots \;\wedge\; Z_{k-1} \synparentof Z_k) \,\rightarrow\, \neg (Z_k\synparentof Z_0)$} \\
    RH1 & \multicolumn{3}{@{}l}{$\vdash \int{\opint{\vec{X}}{\vec{x}}}(\gamma_1\wedge\gamma_2) \,\leftrightarrow\, (\int{\opint{\vec{X}}{\vec{x}}}\gamma_1 \wedge \int{\opint{\vec{X}}{\vec{x}}}\gamma_2)$} \\
    RH2 & \multicolumn{3}{@{}l}{$\vdash \int{\opint{\vec{X}}{\vec{x}}}\neg\gamma \,\leftrightarrow\, \neg\int{\opint{\vec{X}}{\vec{x}}}\gamma$} \\
    EX  & \multicolumn{3}{@{}l}{$\vdash U{=}u \leftrightarrow \int{\opint{\vec{X}}{\vec{x}}}U{=}u$ \qquad for $U \in \XV$ with $U \notin\vec{X}$} \\
    \midrule
    \multicolumn{4}{l}{\textit{Epistemic}:} \\
    K & $\vdash K(\phi\rightarrow\psi)\rightarrow(K\phi\rightarrow K\psi)$ &
    T & $\vdash K\phi\rightarrow\phi$ \\
    N & From $\vdash \phi$ derive $\vdash K\phi$ &
    4 & $\vdash K\phi\rightarrow KK\phi$ \\
    \multicolumn{2}{l}{} &
    5 & $\vdash \neg K\phi\rightarrow K\neg K\phi$ \\
    \midrule
    \multicolumn{4}{l}{\textit{Epistemic+Intervention}:} \\
    CM & \multicolumn{3}{@{}l}{$\vdash \int{\opint{\vec{X}}{\vec{x}}}K\gamma \,\leftrightarrow\, K\int{\opint{\vec{X}}{\vec{x}}}\gamma$} \\
    \midrule
    \multicolumn{4}{l}{\textit{Announcement}:} \\
    RP1 & $\vdash [\psi!]\int{\opint{\vec{X}}{\vec{x}}}Z{=}z \,\leftrightarrow\, (\psi \rightarrow \int{\opint{\vec{X}}{\vec{x}}}Z{=}z)$ &
    RP3 & $\vdash [\psi!](\phi \wedge \chi) \,\leftrightarrow\, ([\psi!]\phi \wedge [\psi!]\chi)$ \\
    RP2 & $\vdash [\psi!]\neg \phi \,\leftrightarrow\, (\psi \rightarrow \neg[\psi!]\phi)$ &
    RP4 & $\vdash [\psi!]K\phi \,\leftrightarrow\, (\psi \rightarrow K(\psi\rightarrow [\psi!]\phi))$ \\
    RE  & \multicolumn{3}{@{}l}{\begin{minipage}[t]{0.8\textwidth}
           From $\vdash \psi_1 \leftrightarrow \psi_2$ derive $\vdash \phi \leftrightarrow \phi[\psi_2/\psi_1]$, with $\phi[\psi_2/\psi_1]$ an \LAfull formula obtained by replacing one or more non-announcement occurrences of $\psi_1$ in $\phi$ with $\psi_2$.\footnotemark\smallskip
         \end{minipage}} \\
    \bottomrule
  \end{footnotesizectabular}
  \caption{Axiom system \LOfull}
  \label{tbl:axioms}
\end{table}

\footnotetext{A non-announcement occurrence of $\psi$ in $\phi$ is an occurrence of $\psi$ in $\phi$ where $\psi$ is not inside the \emph{brackets} of an announcement operator.}

{\medskip}

The axiom system \LOfull is sound and complete for \LAfull in epistemic causal models. Here is the argument for soundness.

\begin{Theorem}
  The axiom system \LOfull is sound for \LAfull in epistemic causal models.
\end{Theorem}
\begin{ourproof}
  For the soundness of HP1-HP6, RH1 and RH2 on causal models (enough for soundness on epistemic causal models, as evaluating the formulas does not require a change in valuation), see \cite{halpern2000axiomatizing}. For the soundness of K, N, T, 4, and 5 on relational structures with an equivalence relation (equivalent to having a simple set of epistemic alternatives, as epistemic causal models have), see \cite{RAK1995,BlackburnRijkeVenema2001}. For the soundness of RP1-RP4 when $[\psi!]$ describes the effect of a deterministic domain-reducing model operation, see \cite{wang2013axiomatizations}.

  For axioms EX and CM, take any $(\tuple{\S, \F, \T}, \A)$. For EX note how, for any $\vec{X}{=}\vec{x}$, the valuations $\A$ and $\newval{\A}{\F}{\opint{\vec{X}}{\vec{x}}}$ assign the same value to \emph{exogenous} variables not occurring in $\vec{X}$ (Definition \ref{def:int:strict}). For CM, note how \begin{inlineenum} \item $K\int{\opint{\vec{X}}{\vec{x}}}\phi$ holds at $(\tuple{\S, \F, \T}, \A)$ iff $\phi$ holds at $(\tuple{\S, \F_{\opint{\vec{X}}{\vec{x}}}, \newval{\T}{\F}{\opint{\vec{X}}{\vec{x}}}}, \newval{\A'}{\F}{\opint{\vec{X}}{\vec{x}}})$ for every $\A' \in \T$, and \item $ \int{\opint{\vec{X}}{\vec{x}}}K\phi$ holds at $(\tuple{\S, \F, \T}, \A)$ iff $\phi$ holds at $(\tuple{\S, \F_{\opint{\vec{X}}{\vec{x}}}, \newval{\T}{\F}{\opint{\vec{X}}{\vec{x}}}}, (\newval{\A}{\F}{\opint{\vec{X}}{\vec{x}}})')$ for every $(\newval{\A}{\F}{\opint{\vec{X}}{\vec{x}}})' \in \newval{\T}{\F}{\opint{\vec{X}}{\vec{x}}}$\end{inlineenum}. Then it is enough to notice how, by Definition \ref{def:intervention2}, the set of relevant valuations for the second, $\newval{\T}{\F}{\opint{\vec{X}}{\vec{x}}}$, is exactly the set of relevant valuations for the first, $\set{\newval{\A'}{\F}{\opint{\vec{X}}{\vec{x}}} \mid \A' \in \T}$. Finally, soundness of RE follows from two facts: the truth-value of every formula depends on the truth-value of its subformulas, and model operations (intervention and announcements) produce epistemic causal models. Thus, substituting a non-announcement subformula for a formula that is semantically equivalent in the given class of structures does not affect the final result.
\end{ourproof}

The argument for completeness follows two steps. \begin{inlineenum} \item First, using the reduction axioms technique, we show that \LOfull allow us to translate any formula in \LAfull into a logically equivalent one without public announcements.\footnote{Readers familiar with DEL might have noticed that \LOfull does not have a reduction axiom for nested announcements $[\phi_1!][\phi_2!]\phi$. There are (at least) two strategies for dealing with such formulas. The first follows an `outside-in' approach, reducing two announcements in a row into a single one. This requires an axiom for nested announcements. The second follows an `inside-out' strategy, applying the reduction over the innermost announcement operator in the formula until the operator disappears, and then proceeding to the next. For this, the rule of substitution of equivalents (our rule RE) is enough \cite[Theorem 11]{wang2013axiomatizations}.} \item Then, relying on the canonical model construction for both causal models \cite{halpern2000axiomatizing} and epistemic models \cite[Chapter 3]{RAK1995}, we show that \LOfull is complete for the language without public announcements\end{inlineenum}. 

\begin{Theorem}\label{thm:completeness}
  The axiom system \LOfull is complete for \LAfull in epistemic causal models.
\end{Theorem}
\begin{ourproof}
  See Appendix \ref{proof:thm:completeness}
\end{ourproof}




\section{Discussion}\label{sec:discussion}

In this section we will compare our proposal to the Causal Team Semantics developed in \cite{barbero2019interventionist,BarSan2020,BarYan2020}. Causal Team Semantics was proposed with the intention of supporting languages that discuss both accidental and causal dependencies. This is a topic that has gained quite some interest in recent years (see, e.g., \cite{ChoHal2004,ibeling2020probabilistic}). Causal Team Semantics was developed along the lines of a non-modal tradition of logics of dependence and independence (e.g. \cite{Vaa2007,ManSanSev2011}) by extending the so-called \emph{team semantics} \cite{Hod1997} with elements taken from causal inference. Even though the focus there  is not on combining causal with epistemic reasoning, this framework bears many similarities to the one we are using, which is why we will discuss it here in detail. Furthermore, this also allows us to say a bit more on the topic of dependence from the perspective of our proposal.

Let us quickly introduce the central notions of Causal Team Semantics to facilitate a comparison of the two frameworks. A causal team\footnote{This is the definition from 
\cite{BarYan2020}, which, save for implementation details,  corresponds to what are called  \emph{fully defined} causal teams in  \cite{barbero2019interventionist} (where a more general notion is considered).} is a tuple $T=\tuple{\T,\F}$ where 
$\mathcal{F}$ is defined similarly as in our paper\footnote{With some additional machinery (which is not worth exploring here) to keep track of the domains of the functions. For simplicity, we may assume here that $\F$ is defined in the same way as for causal epistemic models.} and $\T$ is a possibly empty set of valuations that comply with $\F$. 
Papers on Causal Team Semantics consider a variety of languages. The focus here is the one we shall call $\LAcod$, which is similar to the standard causal language (thus allowing to express various notions of causal dependence in terms of counterfactuals) except for 
the additional \emph{dependence atoms} ``$\dep{X_1,...,X_n}{Y}$'', which expresses (accidental) dependency of the variable $Y$ on the variables $X_1$ to $X_n$. 
A sentence $\dep{X_1,...,X_n}{Y}$ is interpreted as the claim that any two states $s$ and $s'$ that agree on the valuation of the variables $X_1, ..., X_n$ also have to agree on the value they assign to $Y$. 
Let us describe the syntax of $\LAcod$ in more detail.

The signatures used in \cite{barbero2019interventionist} are pairs of the form $\tuple{Dom,Ran}$, where $Dom$ is a set of variables (\emph{not} encoding the distinction between exogenous and endogenous variables) and $Ran$ is defined analogously as the $\mathcal{R}$ used in this paper. For any such fixed signature $\S$, the language $\LAcod$ is defined as  
 \[
    \begin{array}{l@{\;::=\;}l@{\quad}l}
      \alpha & Z{=}z \mid Z{\neq}z \mid \alpha \land \alpha \mid \alpha\lor \alpha \mid  \alpha \supset \alpha \mid \opint{\vec{X}}{\vec{x}}\boxright \alpha
              \\
      \phi   & Z{=}z \mid Z{\neq}z \mid \ \dep{\vec{X}}{Y} \mid \phi \land \phi \mid \phi\lor\phi \mid \alpha\supset\phi \mid \opint{\vec{X}}{\vec{x}}\boxright \phi
    \end{array}
  \]
for $Z,Y,\vec X\in Dom$, $z\in Ran(Z)$, and with the expression $\vec{X}{=}\vec{x}$ an abbreviation for a conjunction of the form $X_1{=}x_1\land\dots\land X_n{=}x_n$.\footnote{One can then have \emph{inconsistent} antecedents, say if $\vec X=\vec x$ contains conjuncts $X=x$ and $X=x'$ with $x\neq x'$. In such cases the intervention is undefined. The semantic clause given in the main text should be extended so as to have any counterfactual with such an antecedent evaluated as (vacuously) true.} \footnote{Notice that the syntax allows negation only at the atomic level. Adding contradictory negation (defined by $T\models \mathord{\sim} \psi$ iff $T\not\models\psi$) would lead to a more expressive language and to an unintended reading of negation. As observed in \cite{BarSan2020}, the language can be extended -- without changes in expressivity -- with a \emph{dual negation}, defined by the clause: $(\T,\F)\models \neg \psi$ iff, for all $s\in\T$,  $({s},\F)\not\models\psi$. The dual negation has the intended reading on formulas without dependence atoms. Neither negation allows the usual interdefinability of $\land$ and $\lor$ via the De Morgan laws; for this reason, both $\land$ and $\lor$ are included in the syntax. } \footnote{Notice also that the antecedent of the operator $\supset$ (selective implication) is restricted to formulas without occurrences of dependence atoms. The consequents of counterfactuals, instead have no restrictions, and they may contain occurrences of $\boxright$.} Below the complete semantics of $\LAcod$  
is given, using the notation of this manuscript. 
Notice that formulas are evaluated on a causal team \emph{globally}: no valuation in $\T$ is isolated as being `the actual world'. At the atomic level, this is done by means of a \emph{universal} quantification. Indeed, while formulas of the form $Z{=}z$ and $Z{\neq}z$ indicate, semantically, that $Z$'s value is (different from) $z$ in \emph{all valuations} in $\T$, a dependence atom $\dep{X_1,...,X_n}{Y}$ indicates, as stated, that \emph{all pairs} of valuations agreeing on the values of all $X_i$ also agree on the value of $Y$. To keep the global perspective through the rest of the formulas, the interpretation of some connectives ($\lor$ and $\supset$) differs from the traditional one (and, in particular, from that given on epistemic causal models). However, these connectives behave classically if applied to subformulas without occurrences of dependence atoms, and also when $\T$ is a singleton (the quantification plays no relevant role).

\begin{smallctabular}{@{}l@{\quad{iff}\quad}l@{}}
  $T \models Z{=}z$    & $s(Z)=z$ for all $s\in \T$ \\
  $T \models Z{\neq}z$ & $s(Z)\neq z$ for all $s\in \T$ \\
  $T \models \ \dep{X_1,...,X_n}{Y}$ & for all $s,s'\in \T$, if $s(X_i)=s'(X_i)$ for $1 \leqslant i \leqslant n$, then $s(Y)=s'(Y)$ \\
  $T \models \phi \land \psi$ & $T\models \phi$ and $T\models \psi$ \\
  $T \models \phi \lor \psi$  & there are $\T_1 \cup \T_2 = \T$ such that $\tuple{\T_1,\mathcal F}\models \phi$ and $\tuple{\T_2,\mathcal F}\models \psi$ \\
  $T \models \alpha \supset \psi$ & $\tuple{\T^{\alpha},\mathcal F} \models \psi$, for $\T^{\alpha} := \set{ s\in \T \mid (\{s\},\mathcal F)\models \alpha }$ and $\alpha$ without \\
  \multicolumn{1}{l}{} & dependence atoms \\
  $T \models \vec X{=}\vec x\boxright\psi$ & $\tuple{\newval{\T}{\F}{\opint{\vec{X}}{\vec{x}}},\mathcal{F}_{\vec X= \vec x}} \models \psi$, with $\newval{\T}{\F}{\opint{\vec{X}}{\vec{x}}}$ and $\mathcal{F}_{\vec X= \vec x}$ as in Definition \ref{def:intervention2}.
\end{smallctabular}

From their definitions, it is clear that an epistemic causal model and a causal team are identical objects; the only difference is that, for evaluating formulas, the former requires an `actual world'. On the syntactic side, even though the truth clauses of the logical operators differ in various respects, we can find several equivalences. For instance, the notion of dependence from team semantics can be expressed in our formal language as well.\footnote{This has been observed, independently and for languages without causal features, in \cite{EijGatYan2017} and \cite{Bal2016}, in the context of epistemic languages with modalities for the knowledge of values.} Indeed, interpret the object $\T$ of a causal team as the epistemic state of some agent. Then, the statement $Y=y$ of causal team semantics can be understood as a claim about the knowledge of the agent, written in our language as $K(Y=y)$. Building on this translation, we can express that variable $Y$ depends on the variables $\vec X$ as the following claim: for all possible valuations $\vec{x}$ of $\vec{X}$ there is some value $y$ of $Y$ such that the agent knows that if she would observe $\vec X = \vec{x}$, she would know that $Y$ has value $y$.
\[ \bigwedge_{\vec{x} \in \mathcal{R}(\vec{X})} \; \bigvee_{y\in\mathcal{R}(Y)} [(X_1{=}x_1 \wedge \cdots \wedge X_n{=}x_n)!]K(Y{=}y). \]


With this idea in mind we can define 
a translation of the non-nested formulas of $\LAcod$.\footnote{A formula is non-nested if, in every subformula of the form $\vec{X}{=}\vec{x}\boxright \phi$, no $\boxright$ occurs inside $\phi$. Providing a translation for these formulas is sufficient, since every formula of the causal team language is provably equivalent to a non-nested one.} 
Setting aside for a moment the case of the operator $\supset$, and using $A$ to denote the set of all possible valuations for $\XV \cup \NV$, the translation is given by the following clauses.

\begingroup
  \small
  \[
    \renewcommand{\arraystretch}{2}
    \begin{array}{c}
      \begin{array}{r@{\;:=\;}l@{\qquad\qquad}r@{\;:=\;}l}
        tr(Y{=}y)     & K(Y{=}y)         & tr(\phi_1\wedge \phi_2) & tr(\phi_1)\wedge tr_(\phi_2) \\
        tr(Y{\neq}y)  & K(\lnot (Y{=}y)) & tr(\vec{X}{=}\vec{x}\boxright \phi) & [\vec{X}{=}\vec{x}]tr(\phi) \\
      \end{array}
      \\

      \begin{array}{r@{\,:=\,}l}
        tr(\phi\vee\psi) &
        \displaystyle \bigvee_{S\subseteq A}K\big([(\bigvee_{\vec{Y}=\vec{y}\in S}\vec{Y}=\vec{y})!]tr(\phi) \;\wedge\; [(\neg\bigvee_{\vec{Y}=\vec{y}\in S}\vec{Y}=\vec{y})!]tr(\psi)\big)\\
       tr(\dep{X_1,...,X_n}{Y})   & \displaystyle \bigwedge_{\vec{x} \in \mathcal{R}(\vec{X})} \; \bigvee_{y\in\mathcal{R}(Y)} [(X_1{=}x_1 \wedge \cdots \wedge X_n{=}x_n)!]K(Y{=}y) \\
      \end{array}
    \end{array}
  \]
\endgroup


A short note on the not-so-intuitive translation clause for $\lor$. First observe that, in the semantic clause for $\lor$, the sets $\T_1$ and $\T_2$ can equivalently be required to form a \emph{partition} of $\T$, i.e. to be disjoint. The translation clause  uses the fact that a partition of $\T$ (say, $\T\cap S$ and $\T\setminus S$) can be characterized by the pair of formulas $\bigvee_{\vec{Y}{=}\vec{y}\in S}\vec{Y}{=}\vec{y}$ (defining $\T\cap S$ as a subset of $\T$) and $\neg\bigvee_{\vec{Y}{=}\vec{y}\in S}\vec{Y}{=}\vec{y}$ (defining $\T\setminus S$). The conjunction $[(\bigvee_{\vec{Y}{=}\vec{y}\in S}\vec{Y}{=}\vec{y})!]tr(\phi) \;\wedge\; [(\neg\bigvee_{\vec{Y}{=}\vec{y}\in S}\vec{Y}{=}\vec{y})!]tr(\psi)$ then ensures that the current assignment either is in $S$ and satisfies the translation of $\phi$, or it is in $\T\setminus S$ and satisfies the translation of $\psi$. The $K$ operator, placed \emph{after} the disjunction $ \bigvee_{S\subseteq A}$, ensures that, fixing a partition, this property holds for all the assignments (i.e. the partition is not picked out as a function of the assignment). Notice also that this translation clause -- as well as that for dependence atoms -- is well-defined relative to a fixed, finite signature, since the translation uses an enumeration of the variables and of their corresponding allowed values.

Formulas of the form $\alpha \supset\psi$ translate into  public announcement formulas. However, in order to play the role of announcement, $\alpha$ cannot be translated using $tr$, as announcements are evaluated according to the classical meaning. We need instead  a simpler translation $e$ which just replaces logical operators with their counterparts in $\LAfull$ ($X{\neq}x$ is replaced by $\neg (X{=}x)$; $\beta\supset\gamma$ by $\beta\rightarrow\gamma$; $\vec{X}=\vec{x}\boxright \phi$ by $[\vec{X}=\vec{x}] \phi$; $\land$ and $\lor$ are left unaltered, or, more precisely, $\beta\lor\gamma$ is replaced by $\neg(\neg\beta \land \neg \gamma)$). Then we can define $tr$ for $\supset$ as follows:

\[
  tr(\alpha \supset \phi) := [e(\alpha) !]tr(\phi)
\]

This translation satisfies the following (for a proof, see Appendix \ref{proof:props}).

\begin{Prop}[Global translation]\label{causalteamtranslation}
  For any causal team $\tuple{\T, \F}$ over a \emph{finite} signature $\S$ and any formula $\phi\in\LAcod$, we have $\tuple{\T, \F} \models \phi$ if and only if, for all $\A \in \T$, we have $(\tuple{\S, \F, \T}, \A) \models tr(\phi)$.
\end{Prop}

This result compares truth on a causal team with  validity over an epistemic causal model. On the other hand, a different translation of the dependence atom from \cite{EijGatYan2017,Bal2016} suggests an alternative, ``local'' translation. Let $tr^*$ be as $tr$, except for the following clauses (notice the additional $K$ operator in both clauses):

\[
  \renewcommand{\arraystretch}{1.4}
  \begin{array}{r@{\;:=\;}l}
    tr^*(\dep{X_1,...,X_n}{Y})  & \displaystyle \bigwedge_{\vec{x} \in \mathcal{R}(\vec{X})} \bigvee_{y\in\mathcal{R}(Y)}K[(X_1{=}x_1\wedge\cdots\wedge X_n{=}x_n)!]K(Y{=}y) \\
    tr^*(\alpha \supset \phi) & K[e(\alpha) !]tr(\phi)\\
  \end{array}
\]

Now we have the following result (for a proof, see Appendix \ref{proof:props}).

\begin{Prop}[Local translation]\label{localtranslation}
  For any causal team $\tuple{\T, \F}$ over a \emph{finite} signature $\S$ and any formula $\phi\in\LAcod$, we have:
  \begin{enumerate}
    \item If $\tuple{\T, \F} \models \phi$, then, for all $\A \in \T$,  $(\tuple{\S, \F, \T}, \A) \models tr(\phi)$.
    \item If \emph{there is} an $\A \in \T$ such that   $(\tuple{\S, \F, \T}, \A) \models tr(\phi)$, then $\tuple{\T, \F} \models \phi$.
  \end{enumerate}
\end{Prop}

This result shows that, \emph{in the finite case}, $\LAfull$ is at least as expressive as $\LAcod$. 
Despite this, the way the notion of (accidental) dependence is spelled out in the two languages differs in an interesting way. While it is a primitive element in the language of Causal Team Semantics, the way it is definable in our epistemic framework emphasises what we can {\it do} with such a concept of dependence: we can make predictions based on what we observe. Furthermore, it is interesting to notice the similarity between this translation of (accidental) dependence and the way causal dependence is expressed. It is also not defined as a primitive in the language, but can be expressed using counterfactuals, which work based on the concept of intervention. These counterfactuals, in turn, focus on what you can do with causal information: prediction based on intervention.

Based on the counterfactual expression, various notions of causal dependence can be defined. We saw one already in Section~\ref{sec:proposal}, Definition~\ref{def:syntparentof}: $X \synparentof Z$, which expresses that $X$ is a causal parent of $Z$ (if $Z$ is an endogenous variable). The local translation of the notion of dependence from Causal Team Semantics into our framework suggests a different notion of causal dependence. We repeat the local translation below under the name of e-dependence. C-dependence defines the corresponding causal notion.\footnote{The additional $K$ operator in the definition of e-dependence is needed to deal with the fact that information update always checks first whether the information that the information state is updated with is true. This problem disappears in the case of interventions, because the formula you intervene with is {\it made} true in the hypothetical scenario you consider.}

\begin{itemize}
    \item $Y$ {\it e-depends} on $X$ in $(E,\A)$ iff $(E,\A)\models \bigwedge_{x\in \mathcal{R}(X)}\bigvee_{y\in\mathcal{R}(Y)}K([(X=x)!]K(Y=y))$
    \item $Y$ {\it c-depends} on $Y$ in $(E,\A)$ iff   $(E,\A)\models \bigwedge_{x\in \mathcal{R}(X)}\bigvee_{y\in\mathcal{R}(Y)}[X=x]K(Y=y)$
\end{itemize}

Given an epistemic causal model, C-dependence holds between a list of variables $X_1, \ldots, X_n$ and a variable $Y$ if any intervention fixing the value of the variables $X_1, \ldots, X_n$ also determines the value of $Y$ {\it within the epistemic state of the agent}. While this notion is certainly more robust than the notion of e-dependence, it still takes into account the epistemic state of the agent. The less the agent knows about the values of the variables, the more variables she needs to control to make sure that a variable $Y$ is in a particular state. If the agent knows more about the actual causal history of $Y$, she can predict the state of $Y$ already from smaller interventions. These kind of hybrid notions between causal and epistemic dependence that our framework allows to define deserve certainly some attention in future research.

\section{Conclusions}

In this paper we have moved some steps towards the integration of causal and epistemic reasoning, providing an adequate semantics, a language combining interventionist counterfactuals with (dynamic) epistemic operators and a sound and complete system of inference. Our deductive system models the thought of an agent reasoning about the consequences of hypothetical interventions and observations. 
It describes what the agent may deduce from her/his \emph{a priori} pool of knowledge about a system of variables. It is therefore a logic of thought experiments. Going back to Example~\ref{circuit} from the introduction, the approach allows us to account for the inference that Billie is not sure that if the button had been pushed, the sprinkler would have been working. However, the logic is not yet able to also model the second inference discussed in connection with this example: if Billie had pushed the button and saw that that the sprinkler works, then she would have known that the circuit is closed.
In order to account for this kind of reasoning we need to model how an agent may reason about (from her perspective) actual experiments. Things change significantly in such a setting: because of unobserved factors, the agent may fail to predict the outcome of an experiment; yet the outcome may sometimes be recovered from direct observation of the consequences of the experiment. %
The development of a such a framework will involve a more careful distinction between \emph{observable} and \emph{unobservable} variables. The resulting logic must necessarily abandon the right-to-left implication of axiom CM ($\int{\opint{\vec{X}}{\vec{x}}}K\phi \,\rightarrow\, K\int{\opint{\vec{X}}{\vec{x}}}\phi$), which expresses the fact that interventions cannot increase the knowledge of the agent. 

Our framework has many points in common with the earlier causal team semantics, and we provided a translation between the two approaches. For the purpose of modeling causal reasoning, our semantics has the advantage, over causal team semantics, of encoding explicitly a notion of actual state of the world (and in particular, of actual value of variables). Actual values seem to be crucial for the attempt of defining notions of \emph{token causation} (\cite{Hit2001b,Woo2003,halpern2016actual}), i.e. causation between events. In order to fully appreciate this advantage, though, we will need to consider richer languages with hybrid features that allow to explicitly refer to the actual values of variables.

Finally, in future work we plan to extend the setting to a multi-agent system. This involves considering not only different agents with potentially different knowledge, but also epistemic attitudes for groups (e.g., distributed and common knowledge) and the effect of inter-agent communication. One advantage this will bring is the potential to contribute to the discussion about causal agency and the role of causation in the study of responsibility within AI (see, for instance, \cite{paper_with_Ilaria}).

\appendix
\section{Appendix}\label{sec:appendix}


\subsection{Proof of Theorem \ref{thm:completeness}}\label{proof:thm:completeness}

As mentioned, the argument for completeness proceeds in two steps: translating any formula in \LAfull into a logically equivalent one without public announcements, and using the canonical model construction for both causal models \cite{halpern2000axiomatizing} and epistemic models \cite{RAK1995} to show that \LOfull is complete for the language without public announcements.

\subsubsection{From \texorpdfstring{\LAfull}{LApakc} to \texorpdfstring{\LAkc}{LAkc}}

The translation of a formula in \LAfull into a logically equivalent one without public announcement operators proceeds in two stages. First, the formula in \LAfull is translated into a logically equivalent one where the only formulas inside the scope of intervention operators are of the form $Z{=}z$. This involves the use of axiom CM for putting epistemic operators $K$ outside the scope of interventions, and the use of axioms RP1-RP4 for eliminating public announcement operators \emph{inside the scope of interventions}. The resulting formula is now built by the free use of Boolean operators, $K$ and $[\psi!]$ over `atoms' of the form $\int{\opint{\vec{X}}{\vec{x}}}Z{=}z$. Then, axioms RP1-RP4 can be applied once more to eliminate \emph{every remaining} public announcement operator.

{\smallskip}

To formalise the process, the following definitions will be useful.

\begin{Def}[Languages $\LA_1$ and $\LAkc$]
  ~
  \begin{itemize}
    \item Formulas $\xi$ of the language $\LA_1$ are given by
    \[
      \begin{array}{l@{\;::=\;}l}
        \xi & \int{\opint{\vec{X}}{\vec{x}}}Z{=}z \mid \lnot \xi \mid \xi \land \xi \mid K\xi \mid [\xi!]\xi
      \end{array}
    \]
    Thus, formulas in $\LA_1$ (a fragment of \LAfull) are built by the free use of Boolean operators, $K$ and $[\psi!]$ over `atoms' of the form $\int{\opint{\vec{X}}{\vec{x}}}Z{=}z$.\footnote{Recall that $Z{=}z$ is the particular case of $\int{\opint{\vec{X}}{\vec{x}}}Z{=}z$ where $\vec{X}$ is empty.}

    \item Formulas $\chi$ of the language \LAkc are given by
    \[
      \begin{array}{l@{\;::=\;}l}
        \chi & \int{\opint{\vec{X}}{\vec{x}}}Z{=}z \mid \lnot \chi \mid \chi \land \chi \mid K\chi
      \end{array}
    \]
    Thus, formulas in \LAkc (a fragment of \LAfull) are then built by the free use of Boolean operators and $K$ over `atoms' of the form $\int{\opint{\vec{X}}{\vec{x}}}Z{=}z$.
  \end{itemize}
\end{Def}

The process consists of two stages: translating from \LAfull into $\LA_1$, and then from $\LA_1$ into \LAkc.

\begin{Prop}
  \begin{inlineenum} \item Every formula $\phi \in \LAfull$ is logically equivalent to a formula $\xi_\phi \in \LA_1$. Moreover, $\phi \leftrightarrow \xi_\phi$ is derivable in \LOfull. \item Every formula $\xi \in \LA_1$ is logically equivalent to a formula $\chi_\xi \in \LAkc$. Moreover, $\xi \leftrightarrow \chi_\xi$ is derivable in \LOfull\end{inlineenum}.
\end{Prop}
\begin{ourproof}
  For \textbf{\textit{(i)}}, consider the translation $\tro:\LAfull \to \LA_1$ given by
  \begingroup
    \small
    \[
      \renewcommand{\arraystretch}{1.4}
      \begin{array}{@{}c@{}c@{}}
        \begin{array}{@{}r@{\,:=\,}l@{}}
          \tro(Z{=}z)                                                   & Z{=}z \\
          \tro(\lnot\phi)                                               & \lnot \tro(\phi) \\
          \tro(\phi_1 \land \phi_2)                                     & \tro(\phi_1) \land \tro(\phi_2) \\
          \tro(K\phi)                                                   & K \tro(\phi) \\
          \tro([\phi'!]\phi)                                            & [\tro(\phi')!]\tro(\phi) \\
        \end{array}
        &
        \begin{array}{@{}r@{\,:=\,}l@{}}
          \tro(\int{\opint{\vec{X}}{\vec{x}}}Z{=}z)                     & \int{\opint{\vec{X}}{\vec{x}}}Z{=}z \\
          \tro(\int{\opint{\vec{X}}{\vec{x}}}\lnot \gamma)              & \tro(\lnot \int{\opint{\vec{X}}{\vec{x}}}\gamma) \\
          \tro(\int{\opint{\vec{X}}{\vec{x}}}(\gamma_1 \land \gamma_2)) & \tro(\int{\opint{\vec{X}}{\vec{x}}}\gamma_1 \land \int{\opint{\vec{X}}{\vec{x}}}\gamma_2) \\
          \tro(\int{\opint{\vec{X}}{\vec{x}}}K\gamma)                   & \tro(K\int{\opint{\vec{X}}{\vec{x}}}\gamma) \\
          \multicolumn{2}{l}{} \\
          \tro(\int{\opint{\vec{X}}{\vec{x}}}[\gamma'!]Z{=}z)           & \tro(\int{\opint{\vec{X}}{\vec{x}}}(\gamma' \rightarrow Z{=}z)) \\
          \tro(\int{\opint{\vec{X}}{\vec{x}}}[\gamma'!]\lnot \gamma)    & \tro(\int{\opint{\vec{X}}{\vec{x}}}(\gamma' \rightarrow \lnot [\gamma'!]\gamma)) \\
          \tro(\int{\opint{\vec{X}}{\vec{x}}}[\gamma'!](\gamma_1 \land \gamma_2)) & \tro(\int{\opint{\vec{X}}{\vec{x}}}([\gamma'!]\gamma_1 \land [\gamma'!]\gamma_2)) \\
          \tro(\int{\opint{\vec{X}}{\vec{x}}}[\gamma'!]K\gamma)         & \tro(\int{\opint{\vec{X}}{\vec{x}}}(\gamma' \rightarrow K(\gamma' \rightarrow [\gamma'!]\gamma))) \\
          \tro(\int{\opint{\vec{X}}{\vec{x}}}[\gamma'!][\gamma''!]\gamma)& \tro(\int{\opint{\vec{X}}{\vec{x}}}[\gamma'!]\tro([\gamma''!]\gamma)) \\
        \end{array}
      \end{array}
    \]
  \endgroup
  From the cases defined in the second column, it should be clear that $\tro$ does yield formulas in $\LA_1$. Indeed, the second and third cases push intervention operators $\int{\opint{\vec{X}}{\vec{x}}}$ through Boolean operators until the formula directly in front of $\int{\opint{\vec{X}}{\vec{x}}}$ is either $Z{=}z$, or else $K$ or else $[\gamma']$. Then, while the fourth case in the second column takes $K$ outside the scope of $\int{\opint{\vec{X}}{\vec{x}}}$, cases six through eight `push' $[\gamma']$ inside the formula until it has only an atom $Z{=}z$ in front, at which moment $[\gamma']$ is eliminated (fifth case).\footnote{Proving that the translation ends and that announcement operators are indeed eventually eliminated requires some care. The crucial thing to notice is that, in cases sixth through eighth, the formula occurring under the scope of announcement operators on the right-hand side is less complex that the one occurring under the scope of the same announcement operator on the left-hand side. See \cite[Section 7.4]{vanDitmarschEtAl2007} and \cite{wang2013axiomatizations} for a detailed explanation of the way the reduction works.} The ninth case deals with nested announcements following an `inside-first` strategy.

  Then, note how $\models \phi \leftrightarrow \tro(\phi)$ holds for every $\phi \in \LAfull$. This can be shown by induction on $\phi$, with the crucial cases being those corresponding to the definitions in the second column. The first is obvious. The second and third follow from the validity of axioms RH1 and RH2, and the fourth follows from CM. Cases fifth through eighth rely on the validity of axioms RP1 through RP4, and the ninth case uses the rule RE. This last rule is used through all the cases, allowing us to replace sub-formulas for logically equivalent ones.

  Finally note how, within the axiom system \LOfull, there is a derivation of $\phi \leftrightarrow \tro(\phi)$, as every non-trivial equivalence that is used for defining the translation (axioms RH1, RH2, CM, RP1-RP4 and rule RE) is in \LOfull.


  {\medskip}

  For \textbf{\textit{(ii)}}, consider the translation $\trt:\LA_1 \to \LAkc$ given by
  \begingroup
    \small
    \[
      \renewcommand{\arraystretch}{1.4}
      \begin{array}{@{}c@{}c@{}}
        \begin{array}{r@{\;:=\;}l}
          \trt(\int{\opint{\vec{X}}{\vec{x}}}Z{=}z) & \int{\opint{\vec{X}}{\vec{x}}}Z{=}z \\
          \trt(\lnot\xi)                            & \lnot \trt(\xi) \\
          \trt(\xi_1 \land \xi_2)                   & \trt(\xi_1) \land \trt(\xi_2) \\
          \trt(K\xi)                                & K \trt(\xi) \\
        \end{array}
        &
        \begin{array}{r@{\;:=\;}l}
          \trt([\xi'!]\int{\opint{\vec{X}}{\vec{x}}}Z{=}z) & \trt(\xi' \rightarrow \int{\opint{\vec{X}}{\vec{x}}}Z{=}z) \\
          \trt([\xi'!]\lnot \xi)                           & \trt(\xi' \rightarrow \neg[\xi'!]\xi) \\
          \trt([\xi'!](\xi_1 \land \xi_2))                 & \trt([\xi']\xi_1 \land [\xi'!]\xi_2) \\
          \trt([\xi'!]K\xi)                                & \trt(\xi' \rightarrow K(\xi' \rightarrow [\xi'!]\xi)) \\
          \trt([\xi'!][\xi''!]\xi)                         & \trt([\xi'!]\trt([\xi''!]\xi)) \\
        \end{array}
      \end{array}
    \]
  \endgroup

  As \cite{wang2013axiomatizations} shows, $\trt$ eliminates public announcement operators, thus yielding indeed a formula in $\LAkc$. Then, note how $\models \xi \leftrightarrow \trt(\xi)$ holds for every $\xi \in \LA_2$. This can be shown by induction on $\chi$: the crucial cases, those corresponding to the definitions in the second column, follow from the validity of axioms RP1 through RP4. For the last entry in the second column, it is the rule RE which allow us to nest the translation function. This last rule is used through all the cases, allowing us to replace sub-formulas for logically equivalent ones. Finally note how, within the axiom system $\LOfull$, there is a derivation $\xi \leftrightarrow \tro(\xi)$, as every non-trivial equivalence defining the translation (axioms RP1-RP4 and rule RE) is in \LOfull.
\end{ourproof}

Then,

\begin{Theorem}
  Every formula $\phi \in \LAfull$ is logically equivalent to a formula $\chi_\phi \in \LAkc$. Moreover, $\phi \leftrightarrow \chi_\phi$ is derivable in \LOfull.
\end{Theorem}

\subsubsection{Canonical model for \texorpdfstring{\LAkc}{LAkc}}

Now it will be shown that \LOkc, the fragment of \LAfull without axioms RH1, RH2, CM and RP1-RP4, is strongly complete for \LAkc over epistemic causal models. This will be done by showing, via the construction of a canonical model, that any \LOkc-consistent set of \LAkc-formulas is satisfiable in a pointed epistemic causal model. The construction here will follow those in \cite{halpern2000axiomatizing} and \cite{RAK1995}, for causal and epistemic models, respectively.

{\medskip}

Let $\mcss$ be the set of all maximally \LOkc-consistent sets of \LAkc-formulas. The first step will be show how each $\Gamma \in \mcss$ gives raise to a causal model.

\begin{Def}[Building a causal model]\label{def:cm}
  Let $\Gamma \in \mcss$ be a maximally \LOkc-consistent set of \LAkc-formulas.
  \begin{itemize}
    \item Let $\vec{U}$ be the tuple of all exogenous variables. For each endogenous variable $V \in \NV$, let $\vec{Y}$ be the tuple of all endogenous variables in $\NV \setminus \set{V}$. The structural function $\fun^{\Gamma}_V$ is defined, for each $\vec{u} \in \R{\vec{U}}$ and $\vec{y} \in \R{\vec{Y}}$, as
    \[
      \fun^{\Gamma}_V(\vec{u}, \vec{y}) = v
      \quad\text{if and only if}\quad
      \int{\opint{\vec{U}}{\vec{u}}, \opint{\vec{Y}}{\vec{y}}}V{=}v \in \Gamma
    \]
    Note: axioms HP1 and HP2 ensure that $\fun^{\Gamma}_V$ is well-defined, as they guarantee $\Gamma$ has one and only one formula of the form $\int{\opint{\vec{U}}{\vec{u}}, \opint{\vec{Y}}{\vec{y}}}V{=}v$ for fixed $\vec{u}$, $\vec{y}$ and $V$. Then, the set of structural functions for $\NV$ in $\Gamma$ defined as $\F^{\Gamma} := \set{\fun^\Gamma_V \mid V \in \NV}$.

    \item The valuation $\A^{\Gamma}$ is defined, for every $Z \in \XV \cup \NV$, as
    \[
      \A^\Gamma(Z) = z
      \quad\text{if and only if}\quad
      Z{=}z \in \Gamma
    \]
    Note: axioms HP1 and HP2 ensure that $\A^{\Gamma}$ is a well-defined function, as they guarantee $\Gamma$ has one and only one formula of the form $Z{=}z$ for a fixed $Z$.
  \end{itemize}
\end{Def}

We show that the structure just defined is indeed a causal model.

\begin{Prop}\label{pro:cm}
  Take $\Gamma \in \mcss$. The tuple $\tuple{\S, \F^{\Gamma}, \A^{\Gamma}}$ is a proper causal model, that is, \begin{inlineenum} \item $\F^{\Gamma}$ is recursive, and \item $\A^\Gamma$ complies with $\F^\Gamma$\end{inlineenum}.
\end{Prop}
\begin{ourproof}
  ~
  \begin{compactenumerate}
    \item Suppose $\F^{\Gamma}$ is not recursive, i.e., suppose $\parentof_{\F^{\Gamma}}^+$ is either not asymmetric or else not transitive. The relation is transitive by construction, so the problem should be asymmetry: there are $X_1 , X_2 \in \XV \cup \NV$ such that $X_1 \parentof_{\F^{\Gamma}}^+ X_2$ and $X_2 \parentof_{\F^{\Gamma}}^+ X_1$, that is,
    \begingroup
      \small
      \[
        X_1 \parentof_{\F^{\Gamma}} Y_1 \parentof_{\F^{\Gamma}} \cdots \parentof_{\F^{\Gamma}} Y_p \parentof_{\F^{\Gamma}} X_2,
        \qquad
        X_2 \parentof_{\F^{\Gamma}} W_1 \parentof_{\F^{\Gamma}} \cdots \parentof_{\F^{\Gamma}} W_q \parentof_{\F^{\Gamma}} X_1
      \]
    \endgroup
    Now, note how, for any two variables $Z_1, Z_2 \in \XV \cup \NV$, if $Z_1 \parentof_{\F^{\Gamma}} Z_2$ then $Z_1 \synparentof Z_2 \in \Gamma$.\footnote{Indeed, let $\vec{Z^-}$ be a vector containing all variables in $(\XV \cup \NV) \setminus \set{Z_1, Z_2}$, and suppose $Z_1 \parentof_{\F^{\Gamma}} Z_2$. By definition of $\parentof_{\F^{\Gamma}}$, there is a vector $\vec{z^-} \in \R{\vec{Z^-}}$ and there are $z_1, z'_1 \in \R{Z_1}$ with $z_1 \neq z'_1$ such that, if $\fun^{\Gamma}_{Z_2}(\vec{z^-}, z_1) = z_2$ and $\fun^{\Gamma}_{Z_2}(\vec{z^-}, z'_1) = z'_2$ (with $\fun^{\Gamma}_{Z_2}$ the structural function for $X_2$ in $\F^\Gamma$), then $z_2 \neq z'_2$. Thus, from the definition of the structural functions in $\F^{\Gamma}$, it follows that $\int{\opint{\vec{Z^-}}{\vec{z^-}}, \opint{Z_1}{z_1}}Z_2{=}z_2 \in \Gamma$ and $\int{\opint{\vec{Z^-}}{\vec{z^-}}, \opint{Z_1}{z'_1}}Z_2{=}z'_2 \in \Gamma$ for $\vec{z^-} \in \R{\vec{Z^-}}$, $z_1 \neq z'_1$ and $z_2 \neq z'_2$. Since $\Gamma$ is maximally consistent, the conjunction of both formulas is also in $\Gamma$, and hence so is $Z_1 \synparentof Z_2$.} Thus, all formulas in
    \begingroup
      \small
      \[
        \left\{
          \begin{array}{l}
            X_1 \synparentof Y_1,\, Y_1 \synparentof Y_2,\, \ldots, Y_{p-1} \synparentof Y_p,\, Y_p \synparentof X_2, \\
            X_2 \synparentof W_1,\, W_1 \synparentof W_2,\, \ldots, W_{q-1} \synparentof W_q,\, W_q \synparentof X_1
          \end{array}
        \right\}
      \]
    \endgroup
    are in $\Gamma$, and so is their conjunction. But, by axiom HP6, $(X_1 \synparentof Y_1 \;\wedge\; \cdots \;\wedge\; W_{q-1} \synparentof W_{q}) \,\rightarrow\, \neg (W_{q}\synparentof X_1) \in \Gamma$. This makes $\Gamma$ inconsistent; a contradiction.

    \item Suppose $\A^\Gamma$ does not comply with $\F^\Gamma$. Then, there is $V \in \NV$ such that $\A^\Gamma(V) = v$ but $\fun^\Gamma_V(\A^\Gamma(\vec{U}), \A^\Gamma(\vec{Y})) \neq v$, with $\vec{U}$ the tuple of all exogenous variables and $\vec{Y}$ the tuple of all endogenous variables in $\NV \setminus \set{V}$. Take $\A^\Gamma(\vec{U}) = \vec{u}$ and $\A^\Gamma(\vec{Y}) = \vec{y}$.

    From $\A^\Gamma$'s definition, $\A^\Gamma(\vec{U}) = \vec{u}$, $\A^\Gamma(\vec{Y}) = \vec{y}$ and $\A^\Gamma(V) = v$ imply that the formulas in $\set{V{=}v} \cup \set{U_i{=}u_i \mid U_i \in \vec{U}} \cup \set{Y_i{=}y_i \mid Y_i \in \vec{Y}}$ are all in $\Gamma$. This and axiom HP3 imply that $\int{\opint{\vec{U}}{\vec{u}}, \opint{\vec{Y}}{\vec{y}}}V{=}v \in \Gamma$. But, from $\fun^\Gamma_V$'s definition, $\fun^\Gamma_V(\A^\Gamma(\vec{U}), \A^\Gamma(\vec{Y})) \neq v$ implies $\int{\opint{\vec{U}}{\vec{u}}, \opint{\vec{Y}}{\vec{y}}}V{=}v \notin \Gamma$, a contradiction.
  \end{compactenumerate}
\end{ourproof}

We have so far been using expressions of the form $\vec X{=} \vec x$ (``assignments'') only inside intervention modalities. From this point onwards we follow the literature and we allow such expressions to occur also outside of modalities; in such contexts, they must be understood as \emph{conjunctions} of atoms, such as $X_1 {=} x_1\land\dots\land X_n{=}x_n$.

\begin{Lemma}[Inverse of composition]\label{lem: inversecomp}
  Let $\vec X,\vec Y, \vec Z$ be tuples of variables in $\XV\cup\NV$, and $\vec x\vec y z\in \R{\vec X\vec Y \vec Z}$. From the assumptions $\int{\opint{\vec{X}}{\vec{x}}}\vec Y{=}\vec y$ and $\int{\opint{\vec{X}}{\vec{x}}, \opint{\vec{Y}}{\vec{y}}}\vec Z{=}\vec z$ we can formally prove $\int{\opint{\vec{X}}{\vec{x}}}\vec Z{=}\vec z$ in \LOkc.
\end{Lemma}

\begin{ourproof}
  Suppose for the sake of contradiction that the set $\Delta= \{[\vec X{=}\vec x]\vec Y{=}\vec y, [\vec X{=}\vec x,\vec Y=\vec y] \vec Z= \vec z\  ,  \neg[\vec X=\vec x] \vec Z= \vec z\}$ is consistent. If $\card(\R{\vec Z})=1$, this contradicts axiom HP2; so assume $\card(\R{\vec Z})>1$. By applying RH2, HP2 and classical logic to the last of these formulas, we obtain that also $\Delta'= \{[\vec X=\vec x]\vec Y=\vec y   ,  [\vec X=\vec x,\vec Y=\vec y] \vec Z=\vec z\  ,  [\vec X=\vec x] \vec Z= \vec z'\}$ is consistent, for some $\vec z'\neq \vec z$. Applying HP3 to the first and third formulas of $\Delta'$, we obtain $[\vec X=\vec x,\vec Y=\vec y] \vec Z= \vec z'$; by HP1 we obtain $\neg[\vec X=\vec x,\vec Y=\vec y] \vec Z= \vec z$, contradicting the consistency of $\Delta'$.
\end{ourproof}

The following proposition is the crucial part of the proof: it shows that $\tuple{\S, \F^\Gamma, \A^\Gamma}$ satisfies all `atoms' (formulas of the form $\int{\opint{\vec{X}}{\vec{x}}}Z{=}z$) in $\Gamma$.

\begin{Prop}\label{pro:truth-lemma-atoms}
  Let $\Gamma \in \mcss$ be a maximally \LOkc-consistent set of \LAkc-formulas. Let $\opint{\vec{X}}{\vec{x}}$ be an assignment, for $\vec{X}$ a tuple of variables in $\XV \cup \NV$; take $Z \in \XV \cup \NV$ and $z \in \R{Z}$. Then,
  \[
    \int{\opint{\vec{X}}{\vec{x}}}Z{=}z \in \Gamma
    \qquad\text{if and only if}\qquad
    \tuple{\S, \F^\Gamma, \A^\Gamma} \models \int{\opint{\vec{X}}{\vec{x}}}Z{=}z
  \]
\end{Prop}
\begin{ourproof}
  From the semantic interpretation, the right-hand side $\tuple{\S, \F^\Gamma, \A^\Gamma} \models \int{\opint{\vec{X}}{\vec{x}}}Z{=}z$ is equivalent to $\newval{\A^\Gamma}{\F^\Gamma}{\opint{\vec{X}}{\vec{x}}}(Z) = z$. Then, the proof will show that, for any assignment $\opint{\vec{X}}{\vec{x}}$ on $\XV \cup \NV$, any $Z \in \XV \cup \NV$ and any $z \in \R{Z}$,
  \[
    \int{\opint{\vec{X}}{\vec{x}}}Z{=}z \in \Gamma
    \qquad\text{if and only if}\qquad
    \newval{\A^\Gamma}{\F^\Gamma}{\opint{\vec{X}}{\vec{x}}}(Z) = z
  \]

  There are two main cases. First, suppose $Z \in \XV$, and take any $\opint{\vec{X}}{\vec{x}}$.
  \begin{compactitemize}
    \item Suppose further that $Z$ occurs in $\vec{X}$, so $Z=X_k$ for some $1 \leqslant k \leqslant \card{\vec{X}}$. \prooflr Suppose $\int{\opint{\vec{X}}{\vec{x}}}X_k{=}z \in \Gamma$. By axiom HP4, we also have $\int{\opint{\vec{X}}{\vec{x}}}X_k{=}x_k \in \Gamma$; thus, axiom HP1 and the consistency of $\Gamma$ imply $z=x_k$. Now, from the definition of the value of intervened variables after an intervention (Definition \ref{def:int:strict}), it follows that $\newval{\A^\Gamma}{\F^\Gamma}{\opint{\vec{X}}{\vec{x}}}(X_k) = x_k$; this, together with $z=x_k$, produces the required $\newval{\A^\Gamma}{\F^\Gamma}{\opint{\vec{X}}{\vec{x}}}(X_k) = z$. \proofrl Suppose $\newval{\A^\Gamma}{\F^\Gamma}{\opint{\vec{X}}{\vec{x}}}(X_k) = z$. From Definition \ref{def:int:strict} again, $\newval{\A^\Gamma}{\F^\Gamma}{\opint{\vec{X}}{\vec{x}}}(X_k) = x_k$, so $z=x_k$. Now, by axiom HP4 again, $\int{\opint{\vec{X}}{\vec{x}}}X_k{=}x_k \in \Gamma$ so, since $z=x_k$, it follows that $\int{\opint{\vec{X}}{\vec{x}}}X_k{=}z \in \Gamma$.

    \item Suppose $Z$ does not occur in $\vec{X}$. By axiom EX, $\int{\opint{\vec{X}}{\vec{x}}}Z{=}z \in \Gamma$ if and only if $Z{=}z \in \Gamma$; by the definition of $\A^\Gamma$ (Definition \ref{def:cm}), $Z{=}z \in \Gamma$ if and only if $\A^\Gamma(Z) = z$; by the definition of the value an intervened valuation assigns to a non-intervened exogenous variable (Definition \ref{def:int:strict}), $\A^\Gamma(Z) = z$ if and only if $\newval{\A^\Gamma}{\F^\Gamma}{\opint{\vec{X}}{\vec{x}}}(Z) = z$.
  \end{compactitemize}

  Suppose now $Z \in \NV$. The proof proceeds by induction on the number of \emph{non-intervened endogenous} variables, i.e., by induction on the size of $\NV \setminus \vec{X}$.
  \begin{inductionproof}
    \item [Case $\bs{\card{\NV \setminus \vec{X}}=0}$.] This is the case when every endogenous variable is being intervened; in particular, $Z$ is. Then, the argument for the case $Z \in \XV$ with $Z$ occurring in $\vec{X}$ shows that the equivalence holds.

    \item [Case $\bs{\card{\NV \setminus \vec{X}}=1}$.] If $Z$ is being intervened (i.e., $Z$ occurs in $\vec{X}$), then the argument for the case $\card{\NV \setminus \vec{X}}=0$ is enough.

    If $Z$ is the lone non-intervened endogenous variable, $\vec{X}$ contains all variables in $\NV \setminus \set{Z}$. Then, define $\opint{\vec{U'}}{\vec{u'}}$ as the assignment over the exogenous variables not in $\vec{X}$ (i.e., $U' \in \vec{U'}$ if and only if both $U' \in \XV$ and $U' \notin \vec{X}$) by taking $u'_i := \A^\Gamma(U'_i)$. From the definition of $\A^\Gamma$, it is clear that $U'_i{=}u'_i \in \Gamma$ for all $U'_i \in \vec{U'}$. Note how the disjoint vectors $\vec{X}$ and $\vec{U'}$ contain, together, exactly all the variables in $(\XV \cup \NV) \setminus \set{Z}$.
    Notice that, by the definition of intervention (Definition \ref{def:int:strict}), we have $\newval{\A^\Gamma}{\F^\Gamma}{\opint{\vec{X}}{\vec{x}}}(Z) = \newval{\A^\Gamma}{\F^\Gamma}{\opint{\vec{X}\vec{U'}}{\vec{x}\vec{u'} }}(Z) = f^{\Gamma}_Z(\vec x,\vec u')$. But then, by the construction of $f^{\Gamma}_Z$ (Definition \ref{def:cm}) we have $\newval{\A^\Gamma}{\F^\Gamma}{\opint{\vec{X}}{\vec{x}}}(Z) = z$ if and only if $[\opint{\vec{X}}{\vec{x}}, \opint{\vec{U'}}{\vec{u'}}]Z{=}z \in \Gamma$. In the presence of $\int{\opint{\vec{X}}{\vec{x}}}\vec{U'}{=}\vec{u'} \in \Gamma$ (a consequence of the previous $\vec U'{=} \vec u'\in \Gamma$ and axiom EX), the latter is equivalent to the required $[\vec X = \vec x]Z = z$ (by Lemma \ref{lem: inversecomp} in one direction, and by axiom HP3 in the other).

    \item [Case $\bs{\card{\NV \setminus \vec{X}} = k > 1}$.] If $Z$ is being intervened, equivalence follows as shown in the case $\card{\NV \setminus \vec{X}}=0$.

    Suppose $Z$ is not being intervened. Define $\opint{\vec{U'}}{\vec{u'}}$ as in the previous case.

    \prooflr Suppose $\int{\opint{\vec{X}}{\vec{x}}}Z{=}z \in \Gamma$. Based on this, we will build a complete valuation $\A^*$, and we will show that $\A^*$ \begin{inlineenum} \item agrees with $\A^\Gamma$ on the values of all exogenous variables not in $\vec{X}$, \item follows $\opint{\vec{X}}{\vec{x}}$ for the values of exogenous variables in $\vec{X}$, and \item complies with all structural functions in $\F^\Gamma_{\vec{X}=\vec{x}}$\end{inlineenum}. Since there is a unique valuation satisfying these three requirements ($\F^\Gamma$ is recursive, as shown in Proposition \ref{pro:cm}), it will follow that $\A^* = \newval{\A^\Gamma}{\F^\Gamma}{\opint{\vec{X}}{\vec{x}}}$. As it will be shown, $\A^*(Z) = z$, so that will produce the required $\newval{\A^\Gamma}{\F^\Gamma}{\opint{\vec{X}}{\vec{x}}}(Z) = z$.

    Recall that $\vec{U'}$ contains exactly all exogenous variables not in $\vec{X}$; let $\vec{V'}$ be the vector containing exactly all \emph{endogenous} variables not in $\vec{X}$. Then, define
    \begin{compactitemize}
      \item $\A^*(X_i) := x_i$ for $X_i \in \vec{X}$;
      \item $\A^*(U'_i) := u'_i$ for $U'_i \in \vec{U'}$;
      \item $\A^*(V'_i) := v'_i$ if and only if $\int{\opint{\vec{X}}{\vec{x}}}V'_i{=}v'_i \in \Gamma$, for $V'_i \in \vec{V'}$.\footnote{Axioms HP1 and HP2 guarantee that this uniquely determines the value of each variable in $\vec{V'}$.}
    \end{compactitemize}

    Note how \begin{inlineenum} \item $\A^*$ agrees with $\A^\Gamma$ on the values of all exogenous variables not in $\vec{X}$ (i.e., variables in $\vec{U'}$) because $\vec{u'}$ is directly taken from $\A^\Gamma$. Moreover, \item it follows $\opint{\vec{X}}{\vec{x}}$ for the values of all (in particular, the exogenous) variables in $\vec{X}$. Then, \item it is only left to show that $\A^*$ complies with $\F^\Gamma_{\vec{X}=\vec{x}}$.\end{inlineenum} For notation, use $y^*$ to denote the value a variable $Y$ receives according to $\A^*$. Note how, since $\card{\NV \setminus \vec{X}} > 1$, there are at least 2 endogenous variables that are not being intervened (i.e., there are at least two variables in $\vec{V'}$); denote them by $W_1$ and $W_2$. By definition of the values in $\vec{v'}$, we have $\int{\opint{\vec{X}}{\vec{x}}}W_1{=}w^*_1 \in \Gamma$ and $\int{\opint{\vec{X}}{\vec{x}}}W_2{=}w^*_2 \in \Gamma$.

    For the proof, it should be shown that, for every \emph{endogenous} variable $Y$, the value $y^*$ complies with the structural function for $Y$ in $\F^\Gamma_{\vec{X}=\vec{x}}$. Take any endogenous variable $Y$ different from $W_1$. If $Y$ is in $\vec{X}$, from axiom HP4 it follows that $\int{\opint{\vec{X}}{\vec{x}}, \opint{W_1}{w^*_1}}Y{=}y^* \in \Gamma$. Otherwise, $Y$ is not in $\vec{X}$, so $Y$ is in $\vec{V'}$ and therefore $\int{\opint{\vec{X}}{\vec{x}}}Y{=}y^* \in \Gamma$. But $\int{\opint{\vec{X}}{\vec{x}}}W_1{=}w^*_1 \in \Gamma$ so, by axiom HP3, $\int{\opint{\vec{X}}{\vec{x}}, \opint{W_1}{w^*_1}}Y{=}y^* \in \Gamma$. Thus, $\int{\opint{\vec{X}}{\vec{x}}, \opint{W_1}{w^*_1}}Y{=}y^* \in \Gamma$ holds for every $Y \in \NV$ different from $W_1$. Since $\card{\NV \setminus (\vec{X} \cup \set{W_1})} = k-1$, from inductive hypothesis it follows that $\newval{\A^\Gamma}{\F^\Gamma}{\opint{\vec{X}}{\vec{x}},\opint{W_1}{w_1^*}}(Y) = y^*$, and also that $\A^*$ complies with the structural function for $Y$ from $\F^\Gamma_{\opint{\vec{X}}{\vec{x}}, \opint{W_1}{w_1^*}}$, since $\A^*$ agrees with $\A^\Gamma$ outside of $\{\vec X, W_1\}$. But $Y$ is different from $W_1$, so $\A^*$ complies with the structural function for $Y$ from $\F^\Gamma_{\opint{\vec{X}}{\vec{x}}}$.

    Thus, for any $Y$ different from $W_1$, the valuation $\A^*$ complies with the structural function for $Y$ at $\F^\Gamma_{\opint{\vec{X}}{\vec{x}}}$. An analogous reasoning shows that, for any $Y$ different from $W_2$, the valuation $\A^*$ complies with the structural function for $Y$ at $\F^\Gamma_{\opint{\vec{X}}{\vec{x}}}$. Thus, for every endogenous variable $Y$, the valuation $\A^*$ complies with the structural function for $Y$ at $\F^\Gamma_{\opint{\vec{X}}{\vec{x}}}$. This proves \textbf{\textit{(iii)}}, so we get the desired $\A^* = \newval{\A^\Gamma}{\F^\Gamma}{\opint{\vec{X}}{\vec{x}}}$. For the final detail, note how our variable $Z$ is in $\vec{V'}$; since we have assumed $\int{\opint{\vec{X}}{\vec{x}}}Z{=}z \in \Gamma$, we have $\A^*(Z) = z$, that is, $\newval{\A^\Gamma}{\F^\Gamma}{\opint{\vec{X}}{\vec{x}}}(Z) = z$, as required.

    {\medskip}

    \proofrl Suppose $\newval{\A^\Gamma}{\F^\Gamma}{\opint{\vec{X}}{\vec{x}}}(Z) = z$. Since $\card{\NV \setminus \vec{X}} = k > 1$, there are at least two endogenous variables not in $\vec{X}$. One of them is $Z$; let $W$ be one of the others, and let $w \in \R{W}$ be the value satisfying $\newval{\A^\Gamma}{\F^\Gamma}{\opint{\vec{X}}{\vec{x}}}(W) = w$.
    \begin{compactitemize}
      \item Consider the valuation $\newval{\A^\Gamma}{\F^\Gamma}{\opint{\vec{X}}{\vec{x}},\opint{W}{w}}$. Since $\newval{\A^\Gamma}{\F^\Gamma}{\opint{\vec{X}}{\vec{x}}}$ and $\newval{\A^\Gamma}{\F^\Gamma}{\opint{\vec{X}}{\vec{x}},\opint{W}{w}}$ agree on $W$, it follows that $\newval{\A^\Gamma}{\F^\Gamma}{\opint{\vec{X}}{\vec{x}},\opint{W}{w}}(Z) = z$. As $\card{\NV \setminus (\vec{X} \cup \set{W})} = k-1$, from the inductive hypothesis it follows that $\int{\opint{\vec{X}}{\vec{x}}, \opint{W}{w}}Z{=}z \in \Gamma$.
      \item Consider the valuation $\newval{\A^\Gamma}{\F^\Gamma}{\opint{\vec{X}}{\vec{x}},\opint{Z}{z}}$. Since $\newval{\A^\Gamma}{\F^\Gamma}{\opint{\vec{X}}{\vec{x}}}$ and $\newval{\A^\Gamma}{\F^\Gamma}{\opint{\vec{X}}{\vec{x}},\opint{Z}{z}}$ agree on $Z$, it follows that $\newval{\A^\Gamma}{\F^\Gamma}{\opint{\vec{X}}{\vec{x}},\opint{Z}{z}}(W) = w$. As $\card{\NV \setminus (\vec{X} \cup \set{Z})} = k-1$, from the inductive hypothesis it follows that $\int{\opint{\vec{X}}{\vec{x}}, \opint{Z}{z}}W{=}w \in \Gamma$.
    \end{compactitemize}
    Thus, $\int{\opint{\vec{X}}{\vec{x}}, \opint{W}{w}}Z{=}z \in \Gamma$ and $\int{\opint{\vec{X}}{\vec{x}}, \opint{Z}{z}}W{=}w \in \Gamma$. Then, by axiom HP5, $\int{\opint{\vec{X}}{\vec{x}}}Z{=}z \in \Gamma$, as required.
  \end{inductionproof}
\end{ourproof}

Having proved this `truth Lemma' for `atoms' in \LAkc, the next step is to go from the causal model $\tuple{\S, \F^\Gamma, \A^\Gamma}$ to an epistemic causal model where all formulas in $\Gamma$ are satisfied. The definition and lemma are below.

\begin{Def}\label{def:ecm}
  Take $\Gamma \in \mcss$.
  \begin{itemize}
    \item Let $\rmcss^\Gamma := \set{ \Gamma' \in \mcss \mid \F^{\Gamma'} = \F^\Gamma}$ be the set maximally consistent sets in $\mcss$ whose structural functions coincide with those of $\Gamma$. Obviously, $\Gamma \in \rmcss^\Gamma$.

    \item Define $R^\Gamma \subseteq \rmcss^\Gamma \times \rmcss^\Gamma$ as $(\Gamma_1, \Gamma_2) \in R^\Gamma$ if and only if $K\chi \in \Gamma_1$ implies $\chi \in \Gamma_2$ for every $\chi \in \LAkc$. This is the standard definition of the relation in modal canonical models (see, e.g., \cite{RAK1995,BlackburnRijkeVenema2001}). The elements of $\rmcss$ are maximally \LOkc-consistent sets, and \LOkc includes axioms T, 4 and 5; thus, it follows from standard modal results (see, e.g., the just mentioned reference) that $R^\Gamma$ is an equivalence relation. In particular, axiom T implies $(\Gamma, \Gamma) \in R^\Gamma$.

    \item Define $\T^{\Gamma} := \set{\A^{\Gamma'} \mid (\Gamma, \Gamma') \in R^\Gamma}$ as containing the valuation function (see Definition \ref{def:cm}) of each maximally consistent set in $\rmcss$ that is $R^\Gamma$-reachable from $\Gamma$. In particular, from $(\Gamma, \Gamma) \in R^\Gamma$ it follows that $\A^\Gamma \in \T^\Gamma$.
  \end{itemize}
  The structure $\ec^\Gamma$ is given by $\tuple{\S, \F^\Gamma, \T^\Gamma}$.
\end{Def}

\begin{Lemma}[Truth lemma for \LAkc]\label{lem:truth-lemma}
  Take $\Gamma \in \mcss$; recall that $\A^\Gamma \in \T^\Gamma$. Then,
  \[ (\tuple{\S, \F^\Gamma, \T^\Gamma}, \A^\Gamma) \models \chi \quad\text{if and only if}\quad \chi \in \Gamma \]
\end{Lemma}
\begin{ourproof}
  The proof is by induction on $\chi \in \Gamma$.
  \begin{inductionproof}
    \item [Case $\bs{\int{\opint{\vec{X}}{\vec{x}}}Z{=}z}$.] The truth-value of an `atom' $\int{\opint{\vec{X}}{\vec{x}}}Z{=}z$ at $(\tuple{\S, \F^\Gamma, \T^\Gamma}, \A^\Gamma)$ is independent from $\T^\Gamma$; then,
    \[
      (\tuple{\S, \F^\Gamma, \T^\Gamma}, \A^\Gamma) \models \int{\opint{\vec{X}}{\vec{x}}}Z{=}z
      \quad\text{if and only if}\quad
      \tuple{\S, \F^\Gamma, \A^\Gamma} \models \int{\opint{\vec{X}}{\vec{x}}}Z{=}z
    \]
    By Proposition \ref{pro:truth-lemma-atoms}, the right-hand side is equivalent to $\int{\opint{\vec{X}}{\vec{x}}}Z{=}z \in \Gamma$.

    \item [Case $\bs{\lnot \chi}$.] Immediate from the inductive hypothesis and the properties of a maximally consistent set.

    \item [Case $\bs{\chi_1 \land \chi_2}$.] Immediate from the inductive hypotheses and the properties of a maximally consistent set.

    \item [Case $\bs{K\chi}$.] As in the same case in the completeness proof of basic modal logic with respect to relational models (see, e.g., \cite[Chapter 3]{RAK1995}), using the fact that \LOkc contains axiom K and rule N.
  \end{inductionproof}
\end{ourproof}

It is only left to check that $\tuple{\S, \F^\Gamma, \T^\Gamma}$ is indeed an epistemic causal model.

\begin{Prop}\label{pro:ecm}
  Take $\Gamma \in \mcss$. The tuple $\tuple{\S, \F^\Gamma, \T^\Gamma}$ is such that every valuation in $\T^\Gamma$ complies with $\F^\Gamma$.
\end{Prop}
\begin{ourproof}
  Take any $\A^{\Gamma'} \in \T^\Gamma$. Note how $\A^{\Gamma'}$ complies with $\F^{\Gamma'}$ (second item in Proposition \ref{pro:cm}). But $\A^{\Gamma'} \in \T^\Gamma$, so $(\Gamma,\Gamma') \in R^\Gamma$ and hence $\Gamma' \in \rmcss$, which implies $\F^{\Gamma'} = \F^\Gamma$. Thus, $\A^{\Gamma'}$ complies with $\F^{\Gamma}$.
\end{ourproof}

Here is, then, the full argument for the strong completeness of \LOkc for \LAkc in epistemic causal models. Let $\Gamma^-$ be any \LOkc-consistent set of \LAkc-formulas. From the enumerability of \LAkc, the set $\Gamma^-$ can be expanded into a maximally \LOkc-consistent set $\Gamma$. By Lemma \ref{lem:truth-lemma}, all formulas in $\Gamma^-$ are satisfiable in $(\tuple{\S, \F^\Gamma, \T^\Gamma}, \A^\Gamma)$, which by Proposition \ref{pro:ecm} is an epistemic causal model.

\subsection{Proofs for Propositions \ref{causalteamtranslation} and \ref{localtranslation}}\label{proof:props}

As pointed out in the main text, here we show how to translate only the \emph{non-nested} formulas of $\LAcod$ 
into $\LAfull$. Furthermore, we denote as $\alpha,\beta,\gamma$...non-nested formulas of $\LAcod$ 
that have no occurrences of dependence atoms. We need to define a simple preliminary translation $e$ of such formulas, so that they may correctly act as public announcements. This will be needed in order to translate formulas of the form $\alpha\supset \psi$.

\begingroup
  \small
  \[
    \renewcommand{\arraystretch}{1.4}
    \begin{array}{@{}r@{\;:=\;}l@{\qquad}r@{\;:=\;}l@{\qquad}r@{\;:=\;}l@{}}
      e(Y{=}y)    & Y{=}y,        & e(\beta\land\gamma) & e(\beta)\land e(\gamma),                 & e(\beta\supset\gamma) & e(\beta)\rightarrow e(\gamma), \\
      e(Y{\neq}y) & \neg (Y{=}y), & e(\beta\lor\gamma)  & \neg(\neg e(\beta)\land \neg e(\gamma)), & e(\vec X = \vec x \boxright \gamma) & [\vec X = \vec x]e(\gamma). \\
    \end{array}
  \]
\endgroup

We point out two simple properties of the preliminary translation $e$.

\begin{Lemma}\label{lem: local correctness of e}
 Let $\alpha$ be a non-nested formula of $\LAcod$ without occurrences of dependence atoms. Let $\tuple{\S,\F,\T}$ be an epistemic causal model. Then, for every $\A\in \T$, 
 \[ (\tuple{\S,\F,\T},\A)\models e(\alpha) \quad\text{if and only if}\quad \tuple{\{\A\},\F}\models \alpha. \]
\end{Lemma}
\begin{ourproof}
    A simple induction on the syntax of $\alpha$.
\end{ourproof}

\begin{Lemma}\label{lem: selection equal on both sides}
 Let $\alpha$ be a non-nested formula of $\LAcod$ 
 without occurrences of dependence atoms. Let $E = \tuple{\S,\F,\T}$ be an epistemic causal model. Then $E^{e(\alpha)} = \tuple{\S,\F,\T^\alpha}$.
\end{Lemma}

\begin{ourproof}
  By definition, $E^{e(\alpha)}$ differs from $E$ only in that its set of valuations is 
  $\{ \A\in\T \mid (E,\A) \models e(\alpha)\}$. But, by Lemma \ref{lem: local correctness of e}, this is equal to $\{ \A\in\T \mid \tuple{\{\A\},\F} \models \alpha\} = \T^{\alpha}$.
\end{ourproof}

Now we can define the translation of (non-nested formulas of) $\LAcod$ into $\LAfull$.
\begingroup
  \small
  \[
    \renewcommand{\arraystretch}{2}
    \begin{array}{c}
      \begin{array}{r@{\;:=\;}l@{\qquad\qquad}r@{\;:=\;}l}
        tr(Y{=}y)     & K(Y{=}y)         & tr(\phi_1\wedge \phi_2) & tr(\phi_1)\wedge tr_(\phi_2) \\
        tr(Y{\neq}y)  & K(\lnot (Y{=}y)) & tr(\vec{X}{=}\vec{x}\boxright \phi) & [\vec{X}{=}\vec{x}]tr(\phi) \\
        \multicolumn{2}{l}{}             & tr(\alpha \supset \phi) & [e(\alpha)!]tr(\phi) \\
      \end{array}
      \\
      \begin{array}{r@{\,:=\,}l}
        tr(\phi\vee\psi) &
        \displaystyle \bigvee_{S\subseteq A}K\big([(\bigvee_{\vec{Y}=\vec{y}\in S}\vec{Y}=\vec{y})!]tr(\phi) \;\wedge\; [(\neg\bigvee_{\vec{Y}=\vec{y}\in S}\vec{Y}=\vec{y})!]tr(\psi)\big)\\
       tr(\dep{X_1,...,X_n}{Y})   & \displaystyle \bigwedge_{\vec{x} \in \mathcal{R}(\vec{X})} \; \bigvee_{y\in\mathcal{R}(Y)} [(X_1{=}x_1 \wedge \cdots \wedge X_n{=}x_n)!]K(Y{=}y) \\
      \end{array}
    \end{array}
  \]
\endgroup
  

We need to show for any causal team $\tuple{\T, \F}$ over a signature $\S$ and any formula $\phi\in\LAcod$, we have $\tuple{\T, \F} \models \phi$ if and only if, for all $\A \in \T$, we have $(\tuple{\S, \F, \T}, \A) \models tr(\phi)$. This can be done by induction on the complexity of $\phi$. We write $E$ for $\tuple{\S, \F, \T}$.

\begin{inductionproof}
  \item [Case $\bs{Y{=}y}$.]
  $\tuple{\T, \F} \models Y{=}y$ iff $\A(Y)=y$ for each $\A\in\T$ iff  $(E, \A) \models K(Y{=}y)$.

  \item [Case $\bs{Y{\neq}y}$.]
  $\tuple{\T, \F} \models Y{\neq}y$ iff $\A(Y)\neq y$ for each $\A\in\T$ iff $(E, \A) \models K(\lnot (Y{\neq}y))$.

  \item [Case $\bs{\psi\wedge\chi}$.]
   This case follows immediately from the inductive hypothesis.

  \item [Case $\bs{\vec{X}{=}\vec{x}\boxright \chi}$.]
  $\tuple{\T, \F} \models \vec{X}{=}\vec{x}\boxright \chi$ iff
  $\tuple{\T^{\F}_{\vec{X}{=}\vec{x}},\F_{\vec{X}{=}\vec{x}}}\models \chi$ iff
  for all $\B\in\T^{\F}_{\vec{X}{=}\vec{x}}$ we have $(\tuple{\S,\F_{\vec{X}{=}\vec{x}},\T^{\F}_{\vec{X}{=}\vec{x}}},\B)\models tr(\chi)$ iff
  for all $\A\in\T$ we have $(\tuple{\S,\F_{\vec{X}{=}\vec{x}},\T^{\F}_{\vec{X}{=}\vec{x}}},\A_{\vec{X}{=}\vec{x}})\models tr(\chi)$ iff
  for all $\A\in\T$ we have $(\tuple{\S,\F,\T},\A)\models [\vec{X}{=}\vec{x}]tr(\chi)$.

 \item [Case $\bs{\alpha\supset\chi}$.]
  $\tuple{\T,\F}\models \alpha\supset \chi$ iff $\tuple{\T^\alpha,\F}\models \chi$ iff (by inductive hypothesis) $(\tuple{\S,\F,\T^\alpha},\A)\models tr(\chi)$ for all $\A\in\T^{\alpha}$ iff (by Lemma \ref{lem: selection equal on both sides}) $(E^{e(\alpha)},\A)\models tr(\chi)$ for all $\A\in\T^{\alpha}$ iff $(E,\A)\models [e(\alpha) !]tr(\chi)$ for all $\A\in\T^{\alpha}$. For the rest, $(E,\A)\models [e(\alpha) !]tr(\chi)$ holds trivially for every $\A\in \T\setminus\T^\alpha$, as $e(\alpha)$ is false on $\A$ by Lemma \ref{lem: local correctness of e}.

  \item [Case $\bs{\psi \lor \chi}$.]
  As a preliminary observation, note how the causal team language is \emph{downward closed}, in the sense that if $\tuple{\T,\F}\models \theta$ and $\T'\subseteq\T$, then $\tuple{\T',\F}\models \theta$ (see \cite{BarSan2020} for a proof). By downward closure, it is easy to see that the statement that there are $\T_1,\T_2$ such that $\T_1\cup \T_2=\T$, $\T_1\models \psi$ and $\T_2\models \chi$ is equivalent to stating the existence of such $\T_1,\T_2$ which are furthermore disjoint.

Now, write $\vec Y$ for $\XV\cup\NV$; recall that $A$ is the set of all possible assignments to $\vec Y$. 
  We have $\tuple{\T, \F} \models \psi\lor\chi$ iff there are \emph{disjoint} $\T_1\cup \T_2 = \T$ such that $\tuple{\T_1,\mathcal F}\models \psi$ and $\tuple{\T_2,\mathcal F}\models \chi$ iff

  (by inductive hypothesis) there are disjoint $\T_1\cup \T_2 = \T$ such that $(\tuple{\S,\F,\T_1},\A)\models tr(\psi)$ for all $\A\in\T_1$ and $(\tuple{\S,\F,\T_2},\A)\models tr(\chi)$ for all $\A\in\T_2$ iff there are disjoint $\T_1\cup \T_2 = \T$ such that $(E,\A)\models [(\bigvee_{\B \in \T_1} \vec Y = \B(\vec y))!]tr(\psi)$ for all $\A\in\T_1$ and $(E,\A)\models [(\bigvee_{\B \in \T_2} \vec Y = \B(\vec y))!]tr(\chi)$ for all $\A\in\T_2$.

  For the next step, notice that the first of these public announcement formulas holds trivially on valuations from $\T_2$ (where the announcement is false); analogously, the second formula holds trivially on valuations from $\T_1$ Thus, the statement above is equivalent to the assertion that both formulas hold on each valuation of $\T$. If furthermore we write $S$ for the set of assignments to $\vec Y$ that correspond to valuations in $\T_1$, since $\T_1$ and $\T_2$ are disjoint we can rewrite the statement as: there is an $S\subseteq A$ such that, for all $\A\in\T$, $(E,\A)\models [(\bigvee_{\vec Y=\vec y \in S} \vec Y = \vec y)!]tr(\psi)\land [(\neg\bigvee_{\vec Y=\vec y \in S} \vec Y = \vec y)!]tr(\chi)$.

  By the semantic clauses, this is equivalent to saying that the same assertion holds for the same formula preceded by $K$. By classical logic, it follows that we can invert the order of the quantifiers, 
  \begin{equation}\label{eqn}
    \begin{array}{l}
      \text{for all } \A\in\T \text{ there is an } S\subseteq A \text{ such that } \\
      (E,\A)\models K\Big([(\bigvee_{\vec Y=\vec y \in S} \vec Y = \vec y)!]tr(\psi) \;\land\; [(\neg\bigvee_{\vec Y=\vec y \in S} \vec Y = \vec y)!]tr(\chi)\Big).
    \end{array}
    \tag{$\ast$}
  \end{equation}

  Then this is equivalent to: for all $\A\in\T$, $(E,\A)\models \bigvee_{S\subseteq A} K\big([(\bigvee_{\vec Y=\vec y \in S} \vec Y = \vec y)!]tr(\psi)\land [(\neg\bigvee_{\vec Y=\vec y \in S} \vec Y = \vec y)!]tr(\chi)\big)$, i.e. the desired conclusion.

  In the opposite direction, 
  assume \eqref{eqn} holds. We need to show that we can swap the two quantifiers; this is not given by a logical rule, but we have instead to show that we can take the same $S$ for all $\A$. But this follows immediately from the clause for $K$: if, for a fixed $S$, we have $(E,\A)\models K\big([(\bigvee_{\vec Y=\vec y \in S} \vec Y = \vec y)!]tr(\psi) \;\land\; [(\neg\bigvee_{\vec Y=\vec y \in S} \vec Y = \vec y)!]tr(\chi)\big)$ for some $\A$, then it holds (with the same $S$) for each $\A\in\T$.


  \item [Case $\bs{\dep{\vec X}{Y}}$.]
  Let $\vec{X}$ be $X_1,\ldots,X_n$. Suppose $\tuple{\T, \F} \models \ \dep{\vec X}{Y}$; this holds iff for all $\A_1, \A_2\in \T$, if $\A_1(X_i)=\A_2(X_i)$ for all $1\leq i\leq n$, then $\A_1(Y)=\A_2(Y)$, that is, iff for all $\vec{x}\in\R{\vec{X}}$ there is some $y\in \R{Y}$ such that for all $\A\in \T$, $\A(X_1)=x_1,...,\A(X_n)=x_n$ implies $\A(Y)=y$, which is equivalent to stating that for all $\vec{x}\in\R{\vec{X}}$, there is some $y\in \R{Y}$ such that for all $\A\in \T$, $(E,\A)\models [(X_1{=}x_1\wedge \cdots \wedge X_n{=}x_n)!]K(Y=y)$. Then, it follows that
  for all $\A\in \T$, $(E,\A)\models\bigwedge_{\vec{x} \in \mathcal{R}(\vec{X})}\bigvee_{y\in\mathcal{R}(Y)}[(X_1{=}x_1\wedge...\wedge X_n{=}x_n)!]K(Y{=}y)$.

  In the opposite direction, supposing that for all $\A\in \T$ the above holds, we only need to prove that $y$ can be chosen independently of $\A$ (i.e., only as a function of $x_1,\dots,x_n$). Actually, we prove that it \emph{must} be chosen independently of $\A$. Suppose for the sake of contradiction that, for some $x_1\dots x_n\in\R{\vec X}$, we have $y\neq y'\in\R{y}$ such that $(\tuple{\S,\F,\T},\A)\models [(X_1{=}x_1\wedge \cdots \wedge X_n{=}x_n)!]K(Y=y)$ and $(\tuple{\S,\F,\T},\A)\models [(X_1{=}x_1\wedge \cdots \wedge X_n{=}x_n)!]K(Y{=}y')$. From this we easily get that every assignment $\B$ in the causal epistemic model $\tuple{\S,\F,\T^{X_1{=}x_1\wedge \cdots \wedge X_n{=}x_n}}$ satisfies both $\B(Y)=y$ and $\B(Y)=y'$, a contradiction.
\end{inductionproof}

{\smallskip}

Similarly, we can prove each of the two claims of Proposition \ref{localtranslation} by induction on the complexity of $\phi$. As before, the case for $\land$ is  trivial. Again, write $E$ for $\tuple{\S, \F, \T}$. In the case of claim \textbf{\textit{(i)}}, for all operators except dependence atoms, we can follow word-by-word the left-to-right entailments from the proof of Proposition \ref{causalteamtranslation}. In the proof of the case $\dep{\vec X}{Y}$ we observe, as an additional step, that from the assumption that for all $\vec{x}\in\R{\vec{X}}$ there is some $y\in \R{Y}$ such that for all $\A \in \T$, $(E,\A) \models [(X_1{=}x_1\wedge\cdots\wedge X_n{=}x_n)!]K(Y{=}y)$ we can infer, by the semantic clause for $K$, that the same statement holds for the formula 
$K[(X_1{=}x_1\wedge\cdots\wedge X_n{=}x_n)!]K(Y=y)$. It is then immediate to conclude that, for all $\A\in \T$, $(E,\A)\models tr^*(\phi)$.

Let us then prove claim \textbf{\textit{(ii)}} of Proposition \ref{localtranslation}.
\begin{inductionproof}
  \item \item [Cases $\bs{X{=}x}$ and $\bs{X{\neq}x}$.]
  Suppose there is $\A\in\T$ such that $(E, \A) \models tr^*(X{=}x)$ (i.e., $K(X{=}x)$). Then, for all $\A\in\T$, $\A(X)=x$, i.e. $\tuple{\T, \F} \models X{=}x$. The proof for $X{\neq}x$ is analogous.

  \item [Case $\bs{\vec{X}{=}\vec{x}\boxright \chi}$.]
  Suppose $(\tuple{\S,\F,\T},\A)\models [\vec{X}{=}\vec{x}]tr^*(\chi)$ holds for some $\A\in\T$. Then, $(\tuple{\S,\F_{\vec{X}{=}\vec{x}},\T^{\F}_{\vec{X}{=}\vec{x}}},\A_{\vec{X}{=}\vec{x}})\models tr^*(\chi)$ and therefore, by inductive hypothesis, $\tuple{\T^{\F}_{\vec{X}{=}\vec{x}},\F_{\vec{X}{=}\vec{x}}}\models \chi$, i.e.,  $\tuple{\T, \F} \models \vec{X}{=}\vec{x}\boxright \chi$.

  \item [Case $\bs{\alpha\supset \chi}$.]
  Suppose there is $\A\in \T$ such that $(E,\A)\models K[e(\alpha)!]tr(\chi)$. Then for all $\A\in \T$ we have $(E,\A)\models [e(\alpha) !]tr(\chi)$. In particular, this holds for all $\A\in \T^\alpha \subseteq \T$, so we can proceed as in the right-to-left direction of the proof of Proposition \ref{causalteamtranslation}.

  \item [Case $\bs{\psi\lor \chi}$.]
  Suppose there is a valuation $\A$ in the set $\T$ satisfying $(E,\A)\models \bigvee_{S\subseteq A}K\big([(\bigvee_{\vec{Y}{=}\vec{y}\in S}\vec{Y}{=}\vec{y})!]tr(\phi)\wedge [(\neg\bigvee_{\vec{Y}{=}\vec{y}\in S}\vec{Y}{=}\vec{y})!]tr(\psi)\big)$. So there is an $S\subseteq A$ such that, for all $\A\in\T$, $(E,\A)\models [(\bigvee_{\vec{Y}=\vec{y}\in S}\vec{Y}=\vec{y})!]tr(\phi)\wedge [(\neg\bigvee_{\vec{Y}=\vec{y}\in S}\vec{Y}=\vec{y})!]tr(\psi)$. From this point we can proceed as in right-to-left direction of the proof of Proposition \ref{causalteamtranslation}.

  \item [Case $\bs{\dep{\vec X}{Y}}$.]
  Suppose there is $\A\in \T$ such that $(E,\A)\models \bigwedge_{\vec{x} \in \mathcal{R}(\vec{X})}\bigvee_{y\in\mathcal{R}(Y)} K[(X_1{=}x_1\wedge\cdots\wedge X_n{=}x_n)!]K(Y{=}y)$; then, for all $x_1,\ldots,x_n\in \mathcal{R}(X_1,\ldots,X_n)$ there is a $y\in\mathcal{R}(Y)$ such that, for all $\A\in \T$, $(E,\A)\models [(X_1{=}x_1\wedge\cdots\wedge X_n{=}x_n)!]K(Y{=}y)$. From this point we can proceed as in the right-to-left case of Proposition \ref{causalteamtranslation}.
\end{inductionproof}

\bibliographystyle{splncs04}
\bibliography{thesis}

\end{document}